\documentclass{article}

\usepackage{microtype}
\usepackage{graphicx}
\usepackage{subfigure}
\usepackage{booktabs} % for professional tables

\usepackage{hyperref}

\usepackage[accepted]{icml2024}

\usepackage{amsmath}
\usepackage{amssymb}
\usepackage{mathtools}
\usepackage{amsthm}

\usepackage[capitalize,noabbrev]{cleveref}

\theoremstyle{plain}

\theoremstyle{definition}

\theoremstyle{remark}

\usepackage[textsize=tiny]{todonotes}

\usepackage{tabularray}
\usepackage{paralist}
\usepackage{makecell}
\usepackage{colortbl}
\usepackage{etoolbox}

\usepackage{rotating}
\usepackage{enumerate}
\usepackage{array}
\usepackage{tabularx}
\usepackage{arydshln}
\usepackage{multirow}
\usepackage{xspace}
\usepackage{utfsym}
\usepackage{bm}
\usepackage{pifont}

\renewcommand{\footnotesize}{\fontsize{8pt}{11pt}\selectfont}

\definecolor{myblue}{RGB}{52,218,247}
\definecolor{myred}{RGB}{255,90,90}
\definecolor{mypink}{RGB}{239,43,159}
\definecolor{myupdate}{RGB}{254,243,222}
\definecolor{myfrozen}{RGB}{237,255,255}
\definecolor{ired}{RGB}{229,72,72}
\definecolor{igreen}{RGB}{80,219,144}
\definecolor{ired}{RGB}{247,142,142}

\definecolor{bluei}{RGB}{218,232,252}

\newcommand{\mlogo}{\raisebox{-6pt}{\includegraphics[width=1.5em]{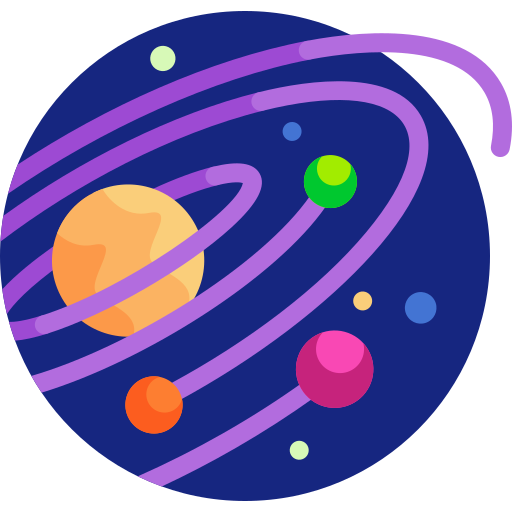}}\xspace\xspace\xspace}
\newcommand{\freeze}{\raisebox{0pt}{\includegraphics[width=0.8em]{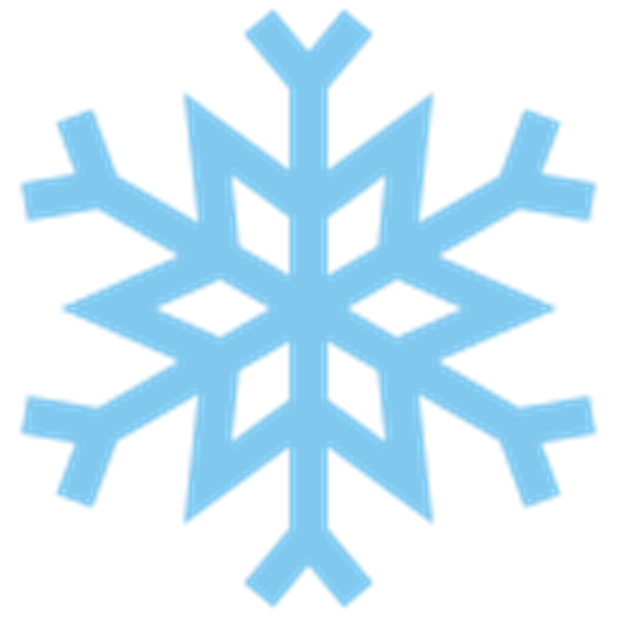}}}
\newcommand{\update}{\raisebox{0pt}{\includegraphics[width=0.8em]{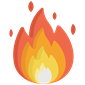}}}

\icmltitlerunning{NExT-GPT: Any-to-Any Multimodal LLM}

\begin{document}

\twocolumn[
\icmltitle{\mlogo {NExT-GPT}: Any-to-Any Multimodal LLM}



\icmlsetsymbol{equal}{*}

\begin{icmlauthorlist}
\icmlauthor{Shengqiong Wu}{yyy}
\icmlauthor{Hao Fei}{yyy}
\icmlauthor{Leigang Qu}{yyy}
\icmlauthor{Wei Ji}{yyy}
\icmlauthor{Tat-Seng Chua}{yyy}
\end{icmlauthorlist}

\icmlaffiliation{yyy}{NExT++ Research Center, National University of Singapore, Singapore}

\icmlcorrespondingauthor{Hao Fei}{haofei37@nus.edu.sg}

\icmlkeywords{Machine Learning, ICML}

\vskip 0.3in
]



\printAffiliationsAndNotice{}  % leave blank if no need to mention equal contribution

\begin{abstract}

While recently Multimodal Large Language Models (MM-LLMs) have made exciting strides, they mostly fall prey to the limitation of only input-side multimodal understanding, without the ability to produce content in multiple modalities. 
As we humans always perceive the world and communicate with people through various modalities, developing any-to-any MM-LLMs capable of accepting and delivering content in any modality becomes essential to human-level AI.
To fill the gap, we present an end-to-end general-purpose any-to-any MM-LLM system, \textbf{NExT-GPT}.
We connect an LLM with multimodal adaptors and different diffusion decoders, enabling NExT-GPT to perceive inputs and generate outputs in arbitrary combinations of text, image, video, and audio. 
By leveraging the existing well-trained high-performing encoders and decoders, NExT-GPT is tuned with only a small amount of parameter (1\%) of certain projection layers, which not only benefits low-cost training but also facilitates convenient expansion to more potential modalities.
Moreover, we introduce a modality-switching instruction tuning (MosIT) and manually curate a high-quality dataset for MosIT, based on which NExT-GPT is empowered with complex cross-modal semantic understanding and content generation.
Overall, our research showcases the promising possibility of building a unified AI agent capable of modeling universal modalities, paving the way for more human-like AI research in the community. 
Project website: \url{https://next-gpt.github.io/}

\end{abstract}

\vspace{-12pt}

\section{Introduction}
\label{Introduction}

\vspace{-1mm}

Recently, the topic of Artificial Intelligence Generated Content (AIGC) has witnessed unprecedented advancements with certain technologies, such as ChatGPT for text generation \citep{chatgpt} and diffusion models for visual generation \citep{abs-2208-13753}. 
Among these, the rise of Large Language Models (LLMs) has been particularly remarkable, e.g., Flan-T5 \citep{abs-2210-11416}, Vicuna \citep{vicuna}, LLaMA \citep{abs-2302-13971} and Alpaca \citep{alpaca}, showcasing their formidable human-level language reasoning and decision-making capabilities, shining a light on the path of Artificial General Intelligence (AGI).
Our world is inherently multimodal, and humans perceive the world with different sensory organs for varied modal information, such as language, images, videos, and sounds, which often complement and synergize with each other.
With such intuition, the purely text-based LLMs have recently been endowed with other modal understanding and perception capabilities of image, video, audio, etc.

A notable approach involves employing adapters that align pre-trained encoders in other modalities to textual LLMs.
This endeavor has led to the rapid development of multimodal LLMs (MM-LLMs), such as BLIP-2 \citep{0008LSH23}, Flamingo \citep{AlayracDLMBHLMM22}, MiniGPT-4 \citep{abs-2304-10592}, Video-LLaMA \citep{abs-2306-02858}, LLaVA \citep{abs-2304-08485}, PandaGPT \citep{abs-2305-16355}, and SpeechGPT \citep{abs-2305-11000}. 
Nevertheless, most of these efforts pay attention to the multimodal content understanding at the input side. 
Lately, fewer works have considered multimodal generation, such as Emu \cite{abs-2307-05222}, \textsc{Dream}LLM \cite{abs-2309-11499}, GILL \cite{abs-2305-17216}, SEED \cite{abs-2307-08041}. 
Notably, these models are confined to generating interleaved texts and images.
We emphasize that natural human cognition and communication indispensably require seamless transitions between any modalities of information.
This makes the exploration of any-to-any MM-LLMs critical, i.e., the ability to accept inputs in any modality and deliver responses in any appropriate modality.

\begin{figure*}[!th]
\centering
\includegraphics[width=0.88\textwidth]{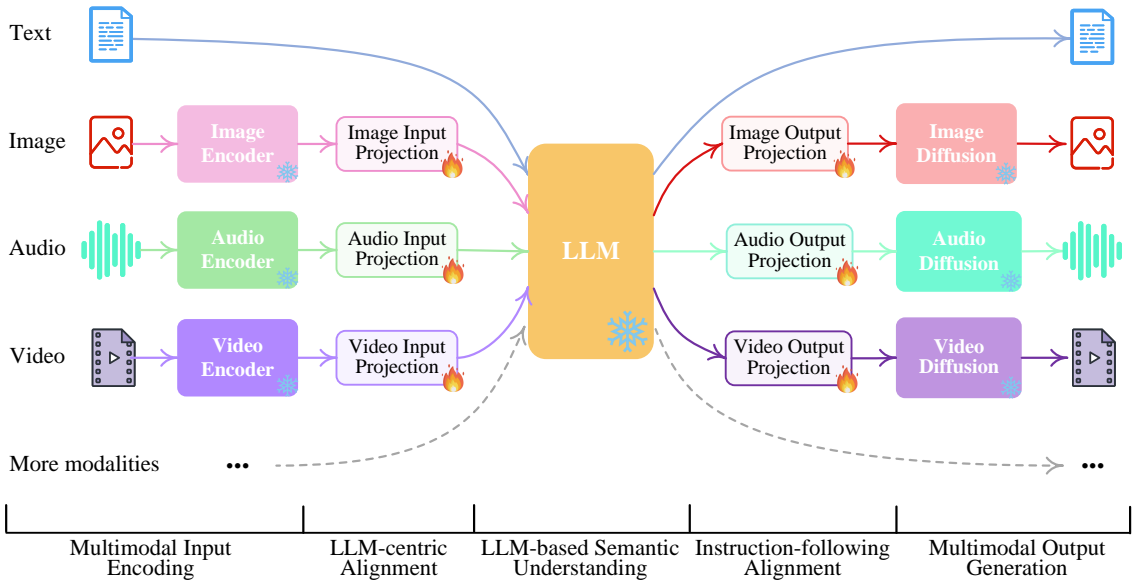}
\caption{
By connecting LLM with multimodal adaptors and diffusion decoders,
NExT-GPT achieves universal multimodal understanding and any-to-any modality input and output. \freeze and \update represent the frozen and trainable modules, respectively.
}
  \vspace{-3mm}
\label{fig:framework}
\end{figure*}

Certain efforts have been made to mimic the human-like any-to-any modality conversion.
Lately, CoDi \citep{abs-2305-11846} has made strides in implementing the capability of simultaneously processing and generating arbitrary combinations of modalities;
however, it lacks the reasoning and decision-making prowess of LLMs as its core, and is also limited to simple paired content generation.
On the other hand, some efforts, e.g., Visual-ChatGPT \citep{abs-2303-04671} and HuggingGPT \citep{abs-2303-17580}, have sought to combine LLMs with external tools to achieve approximately the `any-to-any' multimodal understanding and generation. 
Unfortunately, these systems suffer from critical challenges due to their complete pipeline architecture.
First, the information transfer between different modules is entirely based on discrete texts produced by the LLM, where the cascading process inevitably introduces noise and propagates errors.
More critically, the entire system leverages existing pre-trained tools for inference only. 
Due to the lack of overall end-to-end training, the capabilities of content understanding and multimodal generation can be very limited, especially in interpreting intricate and implicit user instructions.
In a nutshell, there is a compelling need to construct an end-to-end MM-LLM of arbitrary modalities.

In pursuit of this goal, we present \textbf{NExT-GPT}, an any-to-any MM-LLM designed to seamlessly handle input and output in any combination of four modalities: text, image, video, and audio. 
As depicted in Figure \ref{fig:framework}, NExT-GPT comprises three tiers.
\textbf{First}, we leverage established encoders to encode inputs in various modalities, where these representations are projected into language-like representations comprehensible to LLM through a projection layer.
\textbf{Second}, we harness an existing open-sourced LLM as the core to process input information for semantic understanding and reasoning. 
The LLM not only directly generates text tokens but also produces unique `modality signal' tokens that serve as instructions to dictate the decoding layers on whether and what modal content to output correspondingly.
\textbf{Third}, after projection, the produced multimodal signals with specific instructions are routed to different encoders and finally generate content in corresponding modalities.

As NExT-GPT encompasses encoding and generation of various modalities, training the system from scratch would entail substantial costs.
Instead, we take advantage of the existing pre-trained high-performance encoders and decoders,
such as CLIP \citep{RadfordKHRGASAM21}, ImageBind \citep{abs-2305-05665} and the state-of-the-art latent diffusion models \citep{RombachBLEO22,abs-2208-12242,zeroscope,abs-2304-08477,LiuCYMLM0P23,HuangHY0LLYLYZ23}. 
By loading the off-the-shelf parameters, we not only avoid cold-start training but also facilitate the potential growth of more modalities.
For feature alignment across the three tiers, we only consider fine-tuning locally the input projection and output projection layers, with an encoding-side LLM-centric alignment and decoding-side instruction-following alignment, where the minimal computational overhead ensures higher efficiency.
Furthermore, to empower our any-to-any MM-LLM with human-level capabilities in complex cross-modal generation and reasoning, we introduce a \emph{modality-switching instruction tuning}, to equip the system with sophisticated cross-modal semantic understanding and content generation.
To combat the absence of such cross-modal instruction tuning data in the community, we manually collect and annotate a \texttt{MosIT} dataset consisting of 5,000 high-quality samples. 
By employing the LoRA technique \citep{HuSWALWWC22}, we fine-tune the overall NExT-GPT system on instruction tuning data, updating both input and output projection layers and certain LLM parameters.

Overall, this work showcases the promising possibility of developing a more human-like MM-LLM agent capable of modeling universal modalities.
The contributions of this research include:
\begin{compactitem}
    \item We, for the first time, present an end-to-end general-purpose any-to-any MM-LLM, named NExT-GPT, capable of semantic understanding and reasoning and generation of free input and output combinations of text, image, video, and audio.

    \item We introduce lightweight alignment learning techniques, the LLM-centric alignment at the encoding side, and the instruction-following alignment at the decoding side, efficiently requiring only minimal parameter adjustments (only 1\% params) while maintaining highly effective semantic alignment.

    \item We annotate a high-quality modality-switching instruction tuning dataset covering intricate instructions across various modal combinations of text, image, video, and audio, aiding MM-LLM with human-like cross-modal content understanding and reasoning.

\end{compactitem}

\vspace{-2mm}
\section{Related Work}

\vspace{-1mm}
\paragraph{Cross-modal Understanding and Generation}

Our world is replete with multimodal information, wherein we continuously engage in the intricate task of comprehending and producing cross-modal content. 
The AI community correspondingly emerges varied forms of cross-modal learning tasks \citep{ZengZLWCW23,DessiBGRFB23,YangMSLS21,DingYHZZYLZSYT21,LiuCYMLM0P23,DorkenwaldMBRDO21}.
Moreover, to generate high-quality content, a multitude of strong-performing methods have been proposed, such as Transformer \citep{VaswaniSPUJGKP17,ZhangGZBCWWG22,DingYHZZYLZSYT21,GeHYYPJHP22}, GANs \citep{Liu0BZ020,BrockDS19,XuZHZGH018,ZhuP0019}, VAEs \citep{VahdatK20,RazaviOV19}, Flow models \citep{abs-2212-10688,BashiriWLJMDDTS21} and the current state-of-the-art diffusion models \citep{abs-2102-05379,abs-2308-05095,mou2023t2i,abs-2212-05032,RombachBLEO22}. 
In particular, the diffusion-based methods have recently delivered a remarkable performance in a plethora of cross-modal generation tasks, such as DALL-E \citep{RameshPGGVRCS21}, Stable Diffusion \citep{RombachBLEO22}.
While all previous efforts of cross-modal learning are limited to the comprehension of multimodal inputs only, CoDi \citep{abs-2305-11846} lately presents groundbreaking development.
Leveraging the power of diffusion models, CoDi possesses the ability to generate any combination of output modalities, including language, image, video, or audio, from any combination of input modalities in parallel. 
Regrettably, CoDi still falls short of achieving human-like deep reasoning of input content, because it can only deliver parallel cross-modal feeding\&generation without any reasoning and decision-marking capabilities.

\vspace{-4mm}
\paragraph{Multimodal Large Language Models}

LLMs have already made a profound impact and revolution on the entire AI community and beyond \citep{chatgpt,gpt4}, where a series of open-source LLMs have greatly spurred advancement and made contributions to the community \citep{vicuna,abs-2302-13971,abs-2304-10592,abs-2305-01278}. 
Building on top of these LLMs, significant efforts have been made to extend them to deal with multimodal inputs and tasks, leading to the development of MM-LLMs.
On the one hand, most researchers build fundamental MM-LLMs by aligning the well-trained encoders of various modalities to the textual feature space of LLMs to perceive other modal inputs \citep{abs-2302-14045,abs-2304-10592,abs-2205-02655,abs-2305-17216}.
For example, Flamingo \citep{AlayracDLMBHLMM22} uses a cross-attention layer to connect a frozen image encoder to the LLMs. 
BLIP-2 \citep{0008LSH23} employs a Q-Former to translate the input image queries to the LLMs.
There are also various similar practices for building MM-LLMs that are able to understand video (e.g., Video-Chat \citep{abs-2305-06355} and Video-LLaMA \citep{abs-2306-02858}), audio (e.g., SpeechGPT \citep{abs-2305-11000}), etc.
Profoundly, PandaGPT \citep{abs-2305-16355} achieves a comprehensive understanding of six different modalities simultaneously by integrating the multimodal encoder, i.e., ImageBind \citep{abs-2305-05665}.

Nevertheless, these MM-LLMs are all limited to only perceiving multimodal data, without the ability to generate content in arbitrary modalities.
To enable LLMs with both multimodal input and output, some efforts explore employing LLMs as decision-makers, and utilizing existing off-the-shelf multimodal encoders and decoders as tools to execute multimodal input and output, such as Visual-ChatGPT \citep{abs-2303-04671}, HuggingGPT \citep{abs-2303-17580}, and AudioGPT \citep{abs-2304-12995}.
As aforementioned, passing messages between modules with pure texts (i.e., LLM textual instruction) under the discrete pipeline scheme will inevitably introduce noises.
Also, the lack of comprehensive tuning across the whole system significantly limits the efficacy of semantics understanding.
Our work takes the mutual benefits of both the above two types, i.e., learning an any-to-any MM-LLM in an end-to-end manner.

\begin{table*}[!t]
\centering
\fontsize{8}{11}\selectfont
\setlength{\tabcolsep}{3mm}
\caption{
Summary of NExT-GPT system configuration.
Only 1\% of parameters need updating during fine-tuning.
}
\vspace{1mm}
\label{tab:configuration}
\begin{tabular}{lcccccccccc}
\hline
\multirow{2}{*}{ }& \multicolumn{2}{c}{\bf Encoder}& \multicolumn{2}{c}{\bf Input Projection} & \multicolumn{2}{c}{\bf LLM}& \multicolumn{2}{c}{\bf Output Projection}&  \multicolumn{2}{c}{\bf Diffusion} \\ 
\cmidrule(r){2-3} \cmidrule(r){4-5}  \cmidrule(r){6-7}  \cmidrule(r){8-9}  \cmidrule(r){10-11} 
& Name & Param & Name & Param & Name & Param & Name & Param & Name & Param \\
\hline
\multirow{1}{*}{\textbf{Text}}  &	{---}  & {---}  &	{---} & {---}  & \cellcolor{myfrozen}  & \cellcolor{myfrozen}  & {---}  & {---}  &	{---} & {---}	 \\
\multirow{1}{*}{\textbf{Image}}  & \cellcolor{myfrozen}  & \cellcolor{myfrozen} & \cellcolor{myupdate} & \cellcolor{myupdate}  & \cellcolor{myfrozen} Vicuna  & \cellcolor{myfrozen} 7B\freeze& \cellcolor{myupdate}Transformer & \cellcolor{myupdate}31M\update	 &	\cellcolor{myfrozen}SD 	& \cellcolor{myfrozen}1.3B\freeze	 \\
\multirow{1}{*}{\textbf{Audio}} & \cellcolor{myfrozen}& \cellcolor{myfrozen} & \cellcolor{myupdate}  & \cellcolor{myupdate}  & \multicolumn{1}{r}{\cellcolor{myfrozen} \tiny{(LoRA} }&\multicolumn{1}{l}{\cellcolor{myfrozen} \tiny{33M\update)}}&	\cellcolor{myupdate}Transformer& \cellcolor{myupdate}31M\update  &	\cellcolor{myfrozen}AudioLDM	& \cellcolor{myfrozen}975M\freeze \\
\multirow{1}{*}{\textbf{Video}} & \multirow{-3}{*}{\cellcolor{myfrozen}ImageBind} & \multirow{-3}{*}{\cellcolor{myfrozen}1.2B\freeze} & \multirow{-3}{*}{\cellcolor{myupdate}Grouping} & \multirow{-3}{*}{\cellcolor{myupdate}28M\update}  & \cellcolor{myfrozen}&\cellcolor{myfrozen} & \cellcolor{myupdate}Transformer &  \cellcolor{myupdate}32M\update & \cellcolor{myfrozen}Zeroscope	& \cellcolor{myfrozen}1.8B\freeze\\
\hline
\end{tabular}
\vspace{-4mm}
\end{table*}

\section{Overall Architecture}

Figure \ref{fig:framework} presents the schematic overview of the NExT-GPT framework, consisting of three main stages: encoding, LLM understanding and reasoning, and decoding.

\vspace{-3mm}
\paragraph{Multimodal Encoding Stage}

First, we leverage existing well-established models to encode inputs of various modalities.
There are a set of alternatives of encoders for different modalities, e.g., CLIP \citep{RadfordKHRGASAM21}, HuBERT \citep{HsuBTLSM21}.
Here we take advantage of the ImageBind \citep{abs-2305-05665}, which is a unified high-performance encoder across six modalities.
With ImageBind, we are spared from managing many numbers of heterogeneous modal encoders.
Then, via a projection layer, different input representations are mapped into language-like representations that are comprehensible to the LLM.

\vspace{-3mm}
\paragraph{LLM Understanding and Reasoning Stage}
An LLM is used as the core agent of NExT-GPT. 
Technically, we employ the Vicuna (7B-v0) \citep{vicuna}, which is the open-source text-based LLM that is widely used in the existing MM-LLMs \citep{abs-2305-16355,abs-2306-02858}.
LLM takes as input the representations from different modalities and carries out semantic understanding and reasoning over the inputs. 
It outputs: 1) the textual responses directly, and 2) signal tokens of each modality that serve as instructions to dictate the decoding layers on whether to generate multimodal contents and what content to produce if yes.

\begin{figure*}[!t]
\centering
\includegraphics[width=0.98\textwidth]{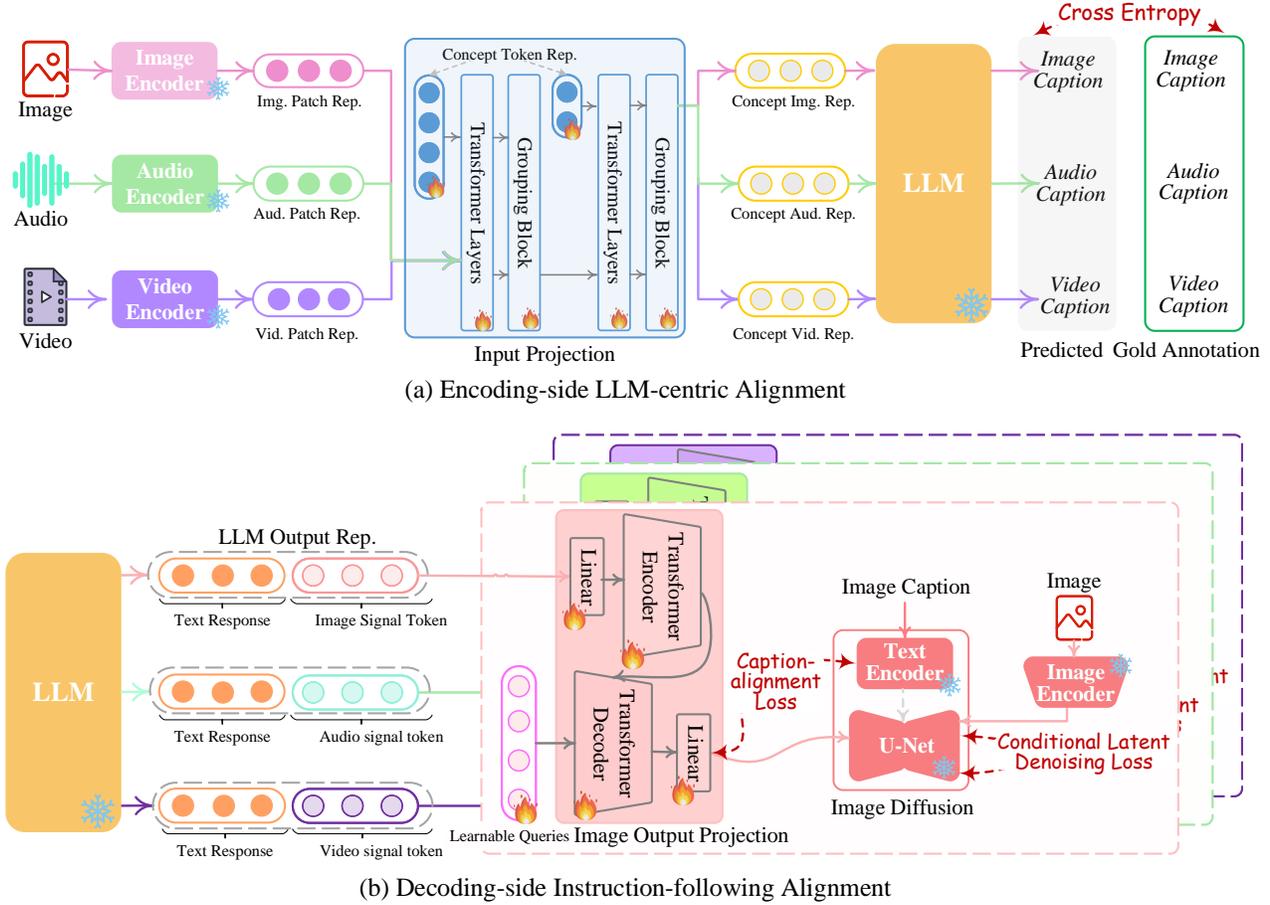}
\vspace{-2mm}
\caption{
Illustration of the lightweight multimodal alignment learning of encoding and decoding, respectively.
}
\vspace{-4mm}
\label{fig:alignment}
\end{figure*}

\vspace{-3mm}
\paragraph{Multimodal Generation Stage}
Receiving the multimodal signals with specific instructions from LLM (if any), the Transformer-based output projection layers map the signal token representations into the ones that are understandable to the following multimodal decoders.
Technically, we employ the current off-the-shelf latent conditioned diffusion models of different modal generations, i.e., Stable Diffusion (\texttt{SD-v1.5}) for image synthesis \citep{RombachBLEO22}, Zeroscope (\texttt{v2-576w}) for video synthesis \citep{zeroscope}, and AudioLDM (\texttt{l-full}) for audio synthesis \citep{LiuCYMLM0P23}.
After a projection layer, the signal representations are fed into the conditioned diffusion models for content generation. 
In Table \ref{tab:configuration} we summarize the overall system configurations.
It is noteworthy that in the entire system, only the input and output projection layers of lower-scale parameters (compared with the overall huge capacity framework) are required to be updated during the following learning, with all the rest of the encoders and decoders frozen.
This amounts to, 155M(=28+33+31+31+32) / [155M + 12.275B(=1.2+7+1.3+1.8+0.975)], or only \textcolor{green}{\bf 1\%} of parameters need to be updated.
This is also one of the key advantages of our MM-LLM.

\vspace{-2mm}
\section{Lightweight Multimodal Alignment Learning}
\label{sec:scene_graph_hallucination}

\vspace{-2mm}
To bridge the gap between the feature space of different modalities, and ensure fluent semantics understanding of different inputs, it is essential to perform alignment learning for NExT-GPT.
Since we design the loosely-coupled system with mainly three tiers, we only need to update the two projection layers at the encoding side and decoding side.

\vspace{-2mm}
\subsection{Encoding-side LLM-centric Multimodal Alignment}
\label{enc-alignment}

\vspace{-2mm}

Most existing MM-LLMs adopt the Transformer-architectured multimodal encoders and generate patch-level grid features (e.g., for image, audio or video).
They transform the multimodal features to be understandable to the core LLM by projecting them into the text feature space straightforwardly via linear layers.
However, we note that the patch-based feature units might not best coincide with the intricate textual token semantics, as intuitively the language tokens always encapsulate separate concepts.
This may result in suboptimal information perception \cite{ZhongYZLCLZDYLG22} in MM-LLMs.
Thus, inspired by \cite{XuMLBBKW22}, we design a type of learnable \emph{concept tokens} to hierarchically aggregate the grid-level features into semantic concept tokens via a grouping mechanism, and then the conceptual representation is fed into LLM.

\vspace{-1mm}
To accomplish the alignment, we adopt an `X-to-text' generation task trained on the `X-caption' pair (`X' stands for image, audio, or video) data from existing corpus and benchmarks, i.e., given the representation of an `X', to prompt the frozen LLM to generate the corresponding text description.
Specifically, we utilize three types of `X-caption' pair data, including 1) `Video-caption' pair dataset: Webvid-2M \citep{BainNVZ21}, a large-scale dataset of short videos with textual description sourced from stock footage sites, 2) `Image-caption' pair dataset: CC3M \citep{SoricutDSG18}, contains over 3 million images accompanied by diverse styles of natural-language descriptions, and 3) `Audio-caption' pair dataset: AudioCaps \citep{KimKLK19}, an extensive dataset of approximately 46k audio clips paired with human-written textual descriptions collected via crowdsourcing.
Figure \ref{fig:alignment}(a) illustrates the learning process.

\vspace{-3mm}
\subsection{Decoding-side Instruction-following Alignment}
\label{dec-alignment}

\vspace{-1mm}
On the decoding end, we have integrated pre-trained conditional diffusion models from external resources. 
Our main purpose is to align the diffusion models with LLM's output instructions. 
However, performing a full-scale alignment process between each diffusion model and the LLM would entail a significant computational burden. 
Alternatively, we explore a more efficient approach, decoding-side instruction-following alignment, as depicted in Figure \ref{fig:alignment}(b).
Specifically, instead of outputting straightforward textual instructions, we design three types of special tokens \citep{abs-2305-17216}, i.e., `\emph{[IMG$_i$]}' ($i=0,\cdots,4$) as image signal tokens; 
`\emph{[AUD$_i$]}' ($i=0,\cdots,8$) as audio signal tokens; and
`\emph{[VID$_i$]}' ($i=0,\cdots,24$) as video signal tokens; these tokens implicitly carry rich and flexible instructions for the downstream diffusion model.
We want to enable the LLM to learn what content to generate, i.e., textual tokens, and modality signal tokens.
If LLM identifies a certain modality content to be produced, a special type of token will be output indicating the activation of that modality; otherwise, no special token output means deactivation of that modality.

We notice that diffusion models generate contents conditioned solely on text-oriented representations, i.e., from the diffusion textual encoders.
However, this text-centered conditioning diverges significantly from the modal signal tokens in our LLM.
This leads to a gap that prevents the diffusion models from accurately interpreting the instructions from LLM.
Thus, on the one hand, we consider taking the LLM's modal signal token representations (after each Transformer-based project layer) as a conditional input in the denoising process to guide the diffusion model to generate appropriate images, videos, or audio.
On the other hand, we also propose minimizing the distance between projected signal token representations and the conditional text representations of the diffusion models to accelerate alignment learning.
Note that all the diffusion backbones (i.e., U-Net) are frozen, which also ensures highly lightweight training.

In the alignment training phase, we take the captions from CC3M, WebVid, and AudioCaps as inputs and concatenate them with the signal tokens as outputs.
The loss function comprises three key components: 1) the negative log-likelihood of producing signal tokens, and 2) the caption alignment loss: $\mathit{l}_2$-distance between the hidden states of signal tokens produced by the LLM and the conditional text representations derived from the text encoder within diffusion models, and 3) conditional latent denoising loss \cite{RombachBLEO22}.

\begin{figure*}[!t]
\centering
\includegraphics[width=0.98\textwidth]{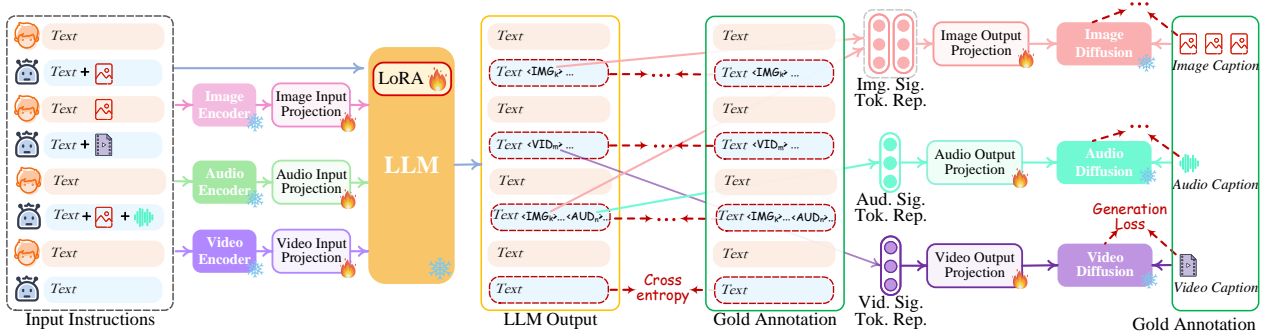}
  \vspace{-2mm}
\caption{
Illustration of modality-switching instruction tuning.
}
\label{fig:instruction_tuning}
\vspace{-4mm}
\end{figure*}

\section{Modality-switching Instruction Tuning}

\vspace{-1mm}
\subsection{Instruction Tuning}

\vspace{-1mm}

Despite aligning both the encoding and decoding ends with LLM, there remains a gap towards the goal of enabling the overall system to faithfully follow and understand users' instructions and generate the desired multimodal outputs. 
To address this, further instruction tuning (IT) \citep{abs-2306-06687,abs-2305-16355,abs-2304-08485} is deemed necessary to enhance the capabilities and controllability of LLM.
IT involves additional training of overall MM-LLMs using `\emph{(INPUT, OUTPUT)}' pairs, where `\emph{INPUT}' represents the user's instruction, and `\emph{OUTPUT}' signifies the desired model output that conforms to the given instruction. 
Technically, we leverage LoRA \citep{HuSWALWWC22} to enable a small subset of parameters within NExT-GPT to be updated concurrently with two layers of projection during the IT phase. 
As illustrated in Figure \ref{fig:instruction_tuning}, when an IT dialogue sample is fed into the system, the LLM reconstructs and generates the textual content of input (and represents the multimodal content with the multimodal signal tokens).
The optimization is imposed based on gold annotations and LLM's outputs.
In addition to LLM tuning, we also fine-tune the decoding end of NExT-GPT. 
We align the modal signal tokens' representation encoded by the output projection with the gold multimodal caption representation encoded by the diffusion condition encoder. 
Thereby, the comprehensive tuning process brings closer to the goal of faithful and effective interaction with users.

\vspace{-2mm}
\subsection{Instruction Dataset}
\label{Instruction Dataset}

\vspace{-1mm}
For the IT of NExT-GPT, we first consider leveraging the well-established `Text'$\to$`Text+X' datasets where `X' could be the image, video, audio, or others, for example, LLaVA-150K \cite{abs-2304-08485}, and VideoChat \cite{abs-2305-06355}.
However, these IT datasets are limited to output textual responses from LLMs.
In our any-to-any scenario, the target not only includes the generations of texts, but also the multimodal contents, i.e., `Text+X'.
Thus, we construct the `Text' $\to$ `Text+X' dataset, i.e., text-to-multimodal (namely T2M) data.
Based on the rich volume of `X-caption' pairs from the existing corpus and benchmarks \citep{SoricutDSG18,LinMBHPRDZ14,BainNVZ21,KimKLK19}, with some templates, we employ GPT-4 to produce varied textual instructions to wrap the captions, and result in the dataset.

\vspace{-3mm}
\paragraph{\texttt{MosIT} Dataset}

Crafting high-quality instructions that comprehensively cover the desired target behaviors is non-trivial. 
We notice that the above IT datasets fail to meet the requirements for our any-to-any MM-LLM scenario.
Firstly, during a human-machine interaction, users and LLM involve diverse and dynamically changing modalities in their inputs and outputs. 
Additionally, we allow multi-turn conversations in the process, and thus the processing and understanding of complex user intentions is required.
However, the above two types of datasets lack variable modalities, and also are relatively short in dialogues, failing to mimic real-world scenarios adequately.

\vspace{-1mm}

To facilitate the development of any-to-any MM-LLM, we propose a novel Modality-switching Instruction Tuning (\texttt{MosIT}) approach. 
\texttt{MosIT} not only supports complex cross-modal understanding and reasoning but also enables sophisticated multimodal content generation.
In conjunction with \texttt{MosIT}, we manually and meticulously construct a high-quality dataset. 
The \texttt{MosIT} dataset encompasses a wide range of multimodal inputs and outputs, offering the necessary complexity and variability to facilitate the training of MM-LLMs that can handle diverse user interactions and deliver the desired responses accurately. 
Specifically, we design some template dialogue examples between a `Human' role and a `Machine' role, based on which we prompt the GPT-4 to generate more conversations under various scenarios with more than 100 topics or keywords.
The interactions are required to be diversified, e.g., including both straightforward and implicit requirements by the `Human', and execution of perception, reasoning, suggestion, and planning, etc., by the `Machine'.
And the interactive content should be logically connected and semantically inherent and complex, with in-depth reasoning details in each response by the `Machine'.
Each conversation should include 3-7 turns (i.e., QA pairs), where the `Human'-`Machine' interactions should involve multiple modalities at either the input or output side, and switch the modalities alternately.
Whenever multimodal contents (e.g., image, audio, and video) are present in the conversations, we look for the best-matched contents from the external resources, including the retrieval systems, e.g., Youtube, and even AIGC tools, e.g., Stable-XL \citep{abs-2307-01952}, and Midjourney.
After human inspections and filtering of inappropriate instances, we obtain a total of 5K high-quality dialogues.
In Table \ref{tab:it-data} of Appendix $\S$\ref{Multimodal IT Datasets Comparison}, we compare the statistics of existing multimodal IT datasets with our \texttt{MosIT} data in detailed statistics.

\begin{table*}[!th]
\centering
\fontsize{8}{11}\selectfont
\setlength{\tabcolsep}{1.5mm}
\caption{
\label{tab:image_perp}
Zero-shot evaluation of image captioning with CIDEr ($\uparrow$) score on NoCaps \cite{AgrawalAD0CJ0BP19}, Flickr 30K \citep{YoungLHH14} and COCO \citep{KarpathyF17}, and image question answering on VQA$^{v2}$ 
 \cite{GoyalKSBP17}, VizWiz \cite{Gurari0SGLGLB18} and OKVQA \cite{MarinoRFM19}, and two evaluation-only benchmarks, MMB  \cite{abs-2307-06281} and SEED \citep{abs-2307-16125}. 
 The best results are marked in bold, and the second ones are underlined.
}
\vspace{1mm}
\begin{tabular}{lccccccccc}
\hline
\bf \multirow{2}{*}{Model} & \multirow{2}{*}{ \bf Version} &  \multicolumn{3}{c}{\bf Image Captioning} & \multicolumn{3}{c}{\bf Image Question Answering} & \multicolumn{2}{c}{\bf Comprehensive} \\
\cmidrule(r){3-5} \cmidrule(r){6-8} \cmidrule(r){9-10}
& & \bf NoCaps & \bf Flickr 30K & \bf COCO & \bf  VQA$^{v2}$  & \bf VizWiz & \bf  OKVQA   & \bf MMB & \bf SEED\\
\hline
InstructBLIP \citep{abs-2305-06500} & Vicuna-7B &  \underline{123.1} & 82.4 & 102.2 & - & 33.4 & 33.9 & 36.0 & -  \\
LLaVA \citep{abs-2304-08485} & LLaMA-2-7B-Chat & 120.7 & \underline{82.7} & - & - & - & -  & 36.2 & -\\
mPLUG-Owl \citep{abs-2304-14178} & LLaMA-7B &  117.0 & 80.3 & \underline{119.3} & - & 39.0 & -  & 46.6 & \underline{34.0}\\
Emu \citep{abs-2307-05222}  & LLaMA-7B   & - & - & 117.7  & 40.0 &  35.4 &  34.7 & - & - \\
DREAMLLM \citep{abs-2309-11499} & Vicuna-7B   & - & - & 115.4 & 56.6 & 45.8  & 44.3 & 49.9 & - \\
Video-LLaVA \citep{abs-2311-10122} & Vicuna-7B & - & - & - & \textbf{74.7} & \underline{48.1} & - & \textbf{60.9} & -\\
\cdashline{1-10}
\rowcolor{bluei} NExT-GPT  & Vicuna-7B &   \bf 123.7 & \bf 84.5 & \bf 124.9  & \underline{66.7} & \textbf{48.4} & \textbf{52.1} & \underline{58.0} & \textbf{57.5}\\
\hline
\end{tabular}
\vspace{-3mm}
\end{table*}

\begin{table*}[!th]
\centering
\fontsize{8}{11}\selectfont
\setlength{\tabcolsep}{1.5mm}
\caption{
\label{tab:video_audio_perp}
Comparison of video reasoning tasks on MSRVTT \citep{XuMYR16}, MSVD-QA and MSRVTT-QA \cite{XuZX0Z0Z17} and NExTQA \cite{XiaoSYC21}, and the audio captioning task on AudioCaps \citep{KimKLK19}. Scores with $*$ means being fine-tuned on the training dataset.
}
\vspace{1mm}
\begin{tabular}{lcccccc}
\hline
\bf \multirow{2}{*}{Model} & \multirow{2}{*}{ \bf Version} &  \bf Video Captioning & \multicolumn{3}{c}{\bf Video Question Answering} & \bf Audio Captioning \\
\cmidrule(r){3-3} \cmidrule(r){4-6} \cmidrule(r){7-7}
& & \bf  MSR-VTT & \bf MSVD-QA & \bf MSRVTT-QA & \bf  NExTQA  & \bf AudioCaps \\
\hline
Codi \citep{abs-2305-11846} & - & \underline{74.4}$^*$ & - & - & - & \underline{78.9}$^*$ \\
UIO-2XXL \citep{abs-2312-17172} & 6.8B & 48.8$^*$ &  41.5 &  52.1 & - & 48.9$^*$  \\
Video-LLaMA \cite{abs-2306-02858} &  LLaMA-7B & - & 51.6 & -  & \underline{29.6} & - \\
Video-LLaVA \citep{abs-2311-10122} & Vicuna-7B & - & \textbf{70.7} & \underline{59.2} & - & -\\
Emu \citep{abs-2307-05222}  & LLaMA-7B   & - & 32.4 & 14.0  &  6.8 &  -  \\
\cdashline{1-7}
\rowcolor{bluei} NExT-GPT  & Vicuna-7B &   \bf \textbf{76.2}$^*$ & \underline{64.5 }& \textbf{61.4 } & \textbf{50.7} & \textbf{81.3}$^*$  \\
\hline
\end{tabular}
\vspace{-3mm}
\end{table*}

\begin{table}[!h]
\centering
\fontsize{8}{11}\selectfont
\setlength{\tabcolsep}{0.9mm}
\vspace{-2mm}
\caption{
\label{tab:syn_avi}
Results on text-to-image/audio/video generation (MS COCO \cite{LinMBHPRDZ14}, AudioCaps \cite{KimKLK19}, and MSRVTT \cite{XuMYR16}). $\dag$: zero-shot results.
}
\vspace{1mm}
\begin{tabular}{lccc}
\hline
\bf \multirow{2}{*}{Model} & \bf Image & \bf Audio & \bf  Video \\
& FID ($\downarrow$)  & FAD ($\downarrow$) &  CLIPSIM ($\uparrow$) \\
\hline
SD-1.5 \citep{WangYFTW22}& 11.21 & - & - \\   
Codi \citep{HuangHY0LLYLYZ23}  & 11.26 & {1.80} & {28.90} \\   
AudioLDM-L \citep{LiuCYMLM0P23}  & - &  1.96 & -\\
GILL-8B$^\dag$ \cite{abs-2305-17216} & 12.20 & -  & - \\
Emu-13B$^\dag$ \cite{abs-2307-05222} & 11.66 & - & - \\
UIO-2XXL \citep{abs-2312-17172} & 13.39 & 2.64 & - \\
\cdashline{1-4}
\rowcolor{bluei} NExT-GPT &   \textbf{10.07} & \textbf{1.68} &  \textbf{31.97} \\ 
NExT-GPT$^\dag$ &   \underline{11.18} & \underline{1.74} &  \underline{30.96} \\ 
\hline
\end{tabular}
\vspace{-5mm}
\end{table}

\vspace{-2mm}
\section{Experiments}

\vspace{-1mm}
In the experiments, we aim to quantify the performance of NExT-GPT on a range of downstream tasks requiring perceiving and generating any modalities.
More settings and implementation details can be found in Appendix $\S$\ref{sec:imp}.

\vspace{-1mm}
Also due to the space limitation, we present a good number of more experimental results and analyses in Appendix $\S$\ref{Additional Experiments}.

\vspace{-1mm}
\subsection{Main Results}
\paragraph{Multimodal Perception}

\vspace{-1mm}
Firstly, we evaluate the semantic understanding capability of the NExT-GPT w.r.t. image, video, or audio, across multiple benchmarks.
The results are shown in Table \ref{tab:image_perp}, and \ref{tab:video_audio_perp}.
Notably, NExT-GPT showcases exceptional performance in image comprehension, demonstrating significant improvements over baseline levels in tasks such as image captioning and image question answering. Moreover, when evaluated on evaluation-only benchmark datasets like MMBench (MMB) and SEED-Bench (SEED), NExT-GPT consistently achieves comparable performance.
Additionally, the model excels in video and audio comprehension. 
In comparison with Codi, NExT-GPT attains enhanced results attributed to its capability for direct text generation from LLM, leveraging the inherent expertise of the LLM.

\vspace{-2mm}
\paragraph{Multimodal Generation}

\vspace{-1mm}
We then examine the synthesis quality of the image, video, or audio conditioned on text. 
Table \ref{tab:syn_avi} presents the comparisons between ours and some state-of-the-art systems.
Overall, NExT-GPT exhibits superior performance in generating images, audio, and video conditioned on text.
Compared to LLM-centric models, i.e., GILL, Emu, and UIO-2XXL, ours stands out by supporting a more diverse range of modalities. 
Moreover, in the generation of individual modalities, NExT-GPT maintains optimal performance, even in zero-shot scenarios.
Notably, in comparison with non-LLM-centered models, ours still demonstrates a clear improvement in generation quality.

\begin{figure*}[!t]
\centering
\includegraphics[width=0.92\textwidth]{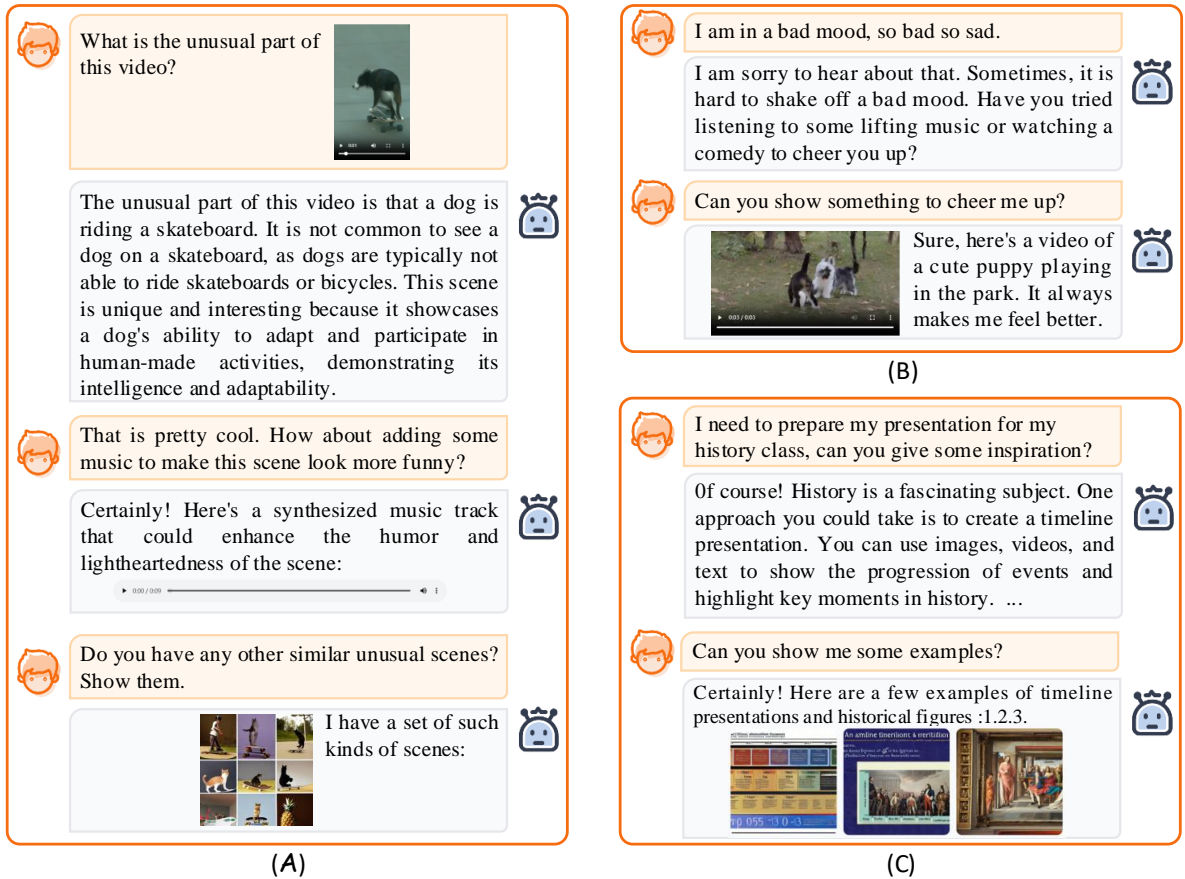}
  \vspace{-3mm}
\caption{
Qualitative examples showcasing the interpretative and generative capabilities of NExT-GPT across diverse modalities or their combinations.
}
\label{fig:demos}
\vspace{-3mm}
\end{figure*}

\begin{figure}[!h]
\centering
\includegraphics[width=0.92\columnwidth]{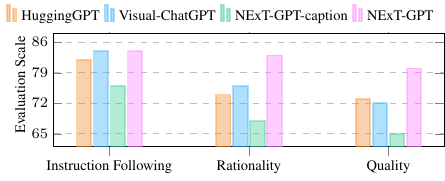}
  \vspace{-4mm}
\caption{
Human Evaluation (1-100 scale, results are on average) of NExT-GPT in comparison with pipeline baselines.
}
\label{fig:human-caption}
\vspace{-4mm}
\end{figure}

\begin{figure*}[!h]
\centering
\includegraphics[width=0.78\textwidth]{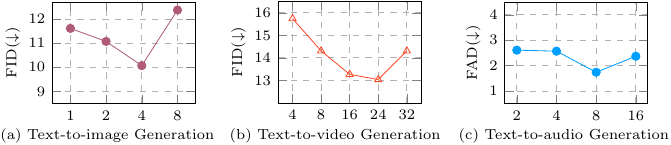}
\vspace{-4mm}
\caption{
The generation quality under different numbers of modality signal tokens.
}
\label{fig:num_signal_tokens}
\end{figure*}

\vspace{-2mm}
\subsection{In-depth Analysis}

\vspace{-1mm}
\paragraph{The Impact of Signal Token Numbers}

In Figure \ref{fig:num_signal_tokens},  we show the results of NExT-GPT utilizing varying numbers of proposed modality-specific signal tokens.
The experimental results reveal that the number of tokens required varies for each modality. Notably, videos, due to their more intricate content generation, demand the highest number of signal tokens. 
The other two modalities, images and audio, achieve satisfactory generation with merely 4 and 8 signal tokens, respectively.
However, the choice of signal token numbers is contingent on factors such as training data size and the selection of the diffusion model. For example, with more extensive data and a robust diffusion model, increasing the number of signal tokens might lead to improved results.

\begin{table*}[!h]
\centering
\fontsize{8}{11}\selectfont
\setlength{\tabcolsep}{4mm}
\caption{
\label{tab:diff_loss}
The perception performance of NExT-GPT by varying input projection mechanisms.
}
\vspace{1mm}
\begin{tabular}{lccccc}
\hline
\bf \multirow{2}{*}{Model} & \multicolumn{2}{c}{Image Question Answering} & \multicolumn{2}{c}{Video Question Answering} & Audio Captioning  \\
\cmidrule(r){2-3} \cmidrule(r){4-5}  \cmidrule(r){6-6}
& VQA$^{v2}$  &  VizWiz & MSVD-QA & MSRVTT-QA & AudioCaps \\
\cdashline{1-6}
\rowcolor{bluei} NExT-GPT & 66.7 & 48.4 & 64.5 & 61.4  & 81.3 \\
\quad w Linear Layer   & 63.8 & 45.4 & 60.8 & 57.1  & 77.4 \\
\quad w Q-former + Linear Layer & 65.1 & 46.9 & 63.4 & 58.1 & 79.7 \\
\hline
\end{tabular}
\vspace{-1mm}
\end{table*}

\vspace{-2mm}

\paragraph{The Impact of Grouping Mechanism}

To further illustrate the effectiveness of employing the grouping mechanism to align visual features with LLM, we conducted experiments with different projection architecture designs. 
These designs include `w Linea Layer' which removes the grouping module and directly maps the output of Imagebind to the language embedding space through a single linear layer, and `w Q-former + Linea Layer' which integrates Q-former instead of the grouping mechanism.
All variants undergo training following the same procedure as the original design. 
The results of two image QA datasets, two video QA datasets, and an audio captioning dataset are presented in Table 4.
The experimental findings indicate a significant decrease in the model's perceptual capabilities across three modalities when using a simple linear approach. 
In addition, the integration of Q-former yields a modest improvement in perceptual capabilities. 
This enhancement might be attributed to the Q-former's ability to perform slight visual feature grouping, aligning effectively with complex textual token semantics, thus elevating perceptual capabilities.
However, our grouping mechanism of NExT-GPT shows the optimal performance.

\vspace{-3mm}
\paragraph{Evaluation on Pipeline vs End-to-End MM-LLMs}

To evaluate if the system really or how well it understands the input and generates output content (response text + image), we perform the human evaluation. For constructing the testing data, we first leverage GPT-4 to synthesize 100 {complex} instructions (e.g., involving intricate and semantically-rich scenes) that require implicit reasoning ability to generate image content. Then, the synthesized instructions are fed into the models to generate the response text + image content. Subsequently, five unbiased volunteers evaluate the generated results under three aspects, 1) \textbf{Instruction following}, identifying, among the four models, which of the generated text+image accurately responded to the input instructions, 2) \textbf{Rationality}, determining which of the generated images adhered to the input instructions, 3) \textbf{Quality}, evaluating which of the generated images exhibited the highest quality.
Figure \ref{fig:human-caption} illustrates superior performance in following complex instructions and generating high-quality images, compared to two existing systems and NExT-GPT-caption, which directly generates textual captions for downstream diffusion models.

\vspace{-4mm}
\paragraph{Qualitative Analysis}
To directly demonstrate the effectiveness and potential of NExT-GPT in developing human-like conversational agents, we further offer qualitative examples that vividly illustrate the system's capacity to comprehend and reason contents across various modalities in any combination, as shown in Figure \ref{fig:demos}.
From example (A), we can note that NExT-GPT can understand the unusual part of the input video, and synthesize a light-heartedness audio and similar unusual scenes, i.e., a cat riding a skateboard.
In addition, beyond responding to explicit queries prompting model synthesis in specific modalities, NExT-GPT demonstrates proficiency in inferring implicit user intentions. 
In example (B), when the user conveys a negative mood, NExT-GPT responds empathetically and autonomously and decides to present a cheerful puppy video to uplift the user's spirits.
Similarly, when preparing a presentation for a history class, NExT-GPT exhibits flexibility in generating pertinent tips and visualizations.
Kindly refer to Appendix $\S$\ref{Example Demonstrations} for more demonstrations with implicit and explicit instructions.

\vspace{-2mm}
\section{Conclusion}

In this work, we presented an end-to-end general-purpose any-to-any multimodal Large Language Model (MM-LLM).
By connecting an LLM with multimodal adaptors and different diffusion decoders, NExT-GPT is capable of perceiving inputs and generating outputs in any combination of text, image, video, and audio. 
Harnessing the existing well-trained highly-performing encoders and decoders, training NExT-GPT only entails a few number of parameters (1\%) of certain projection layers, which not only benefits low costs but also facilitates convenient expansion of more potential modalities in the future.
To enable our NExT-GPT with complex cross-modal semantic understanding and content generation, we further introduced a modality-switching instruction tuning (\texttt{MosIT}), and manually curated a high-quality dataset for MosIT.
Overall, our research showcases the potential of any-to-any MM-LLMs in bridging the gap between various modalities and paving the way for more human-like AI systems in the future.

\section*{Acknowledgements}

This work is supported by CCF-Baidu Open Fund and NExT Research Center.

\section*{Impact Statement}

This paper aims to develop a human-level AI agent, an end-to-end general-purpose any-to-any MM-LLM. 
The NExT-GPT, constrained by the quantity of fine-tuning data and the quality of base models, may produce low-quality or hallucinated content that could be harmful. 
Users are cautioned to interpret results carefully and adhere to licensing rules, with commercial use prohibited. 
We prioritize data privacy by following social media platform terms and obtaining user consent when necessary, ensuring all personal information is anonymized or obfuscated. 
Additionally, we are vigilant in minimizing bias in dataset collection, striving for a representative and fair dataset that does not favor or disfavor any particular group or perspective.

\nocite{langley00}

\bibliography{example_paper}
\bibliographystyle{icml2024}

%%%%%%%%%%%%%%%%%%%%%%%%%%%%%%%%%%%%%%%%%%%%%%%%%%%%%%%%%%%%%%%%%%%%%%%%%%%%%%%
%%%%%%%%%%%%%%%%%%%%%%%%%%%%%%%%%%%%%%%%%%%%%%%%%%%%%%%%%%%%%%%%%%%%%%%%%%%%%%%
% APPENDIX
%%%%%%%%%%%%%%%%%%%%%%%%%%%%%%%%%%%%%%%%%%%%%%%%%%%%%%%%%%%%%%%%%%%%%%%%%%%%%%%
%%%%%%%%%%%%%%%%%%%%%%%%%%%%%%%%%%%%%%%%%%%%%%%%%%%%%%%%%%%%%%%%%%%%%%%%%%%%%%%
\newpage
\appendix
\onecolumn

\section{Potential Limitation and Future work} 

As future work, there are at least following four avenues to explore.
\setdefaultleftmargin{1.9em}{2.2em}{2.87em}{1.7em}{2em}{1em}
\begin{compactitem}

    \item \textbf{i) Modalities \& Tasks Expansion}: Due to resource limitations, currently, our system supports input and output in four modalities: language, images, videos, and audio. Next, we plan to extend this to accommodate even more modalities (e.g., web page, 3D vision, heat map, tables\&figures) and tasks (e.g., object detection, segmentation, grounding, and tracking), broadening the system's applicability to become more universal.

    \item \textbf{ii) LLM Variants}: Currently, we have implemented the 7B Vicuna version of the LLM. Our next plans involve incorporating various LLM types and sizes, allowing practitioners to choose the most suitable one for their specific requirements.

    \item \textbf{iii) Multimodal Generation Strategies}: While our system excels in generating content across modalities, the quality of generative outputs can sometimes be limited by the capabilities of the diffusion model. It is very promising to explore the integration of retrieval-based approaches to complement the generative process, potentially improving the overall system's performance.

    \item \textbf{iv) MosIT Dataset Expansion}: Currently, our IT dataset has room for expansion. We intend to significantly increase the amount of annotated data, ensuring a more comprehensive and diverse set of instructions to further enhance the MM-LLMs' ability to understand and follow user prompts effectively.
\end{compactitem}

\vspace{-2mm}
\section{Full Related Work}

\paragraph{Cross-modal Understanding and Generation}

Our world is replete with multimodal information, wherein we continuously engage in the intricate task of comprehending and producing cross-modal content. 
The AI community correspondingly emerges varied forms of cross-modal learning tasks, such as Image/Video Captioning \citep{ZengZLWCW23,DessiBGRFB23,MilewskiMC20,GuCWZLW23,LinLL0G0LW22}, Image/Video Question Answering \citep{YangMSLS21,XiaoYL0JC22,0004WXJC22,abs-2306-16762,00010BT0GZ18}, Text-to-Image/Video/Speech Synthesis \citep{abs-2209-14792,abs-2205-15868,abs-2303-09522,abs-2208-01618,DingYHZZYLZSYT21,LiuCYMLM0P23,HuangHY0LLYLYZ23}, Image-to-Video Synthesis \citep{DorkenwaldMBRDO21,abs-2304-06025} and more, all of which have experienced rapid advancements in past decades.
Researchers have proposed highly effective multimodal encoders, intending to construct unified representations encompassing various modalities. 
Meanwhile, owing to the distinct feature spaces of different modalities, it is essential to undertake modality alignment learning.
Moreover, to generate high-quality content, a multitude of strong-performing methods have been proposed, such as Transformer \citep{VaswaniSPUJGKP17,ZhangGZBCWWG22,DingYHZZYLZSYT21,GeHYYPJHP22}, GANs \citep{Liu0BZ020,BrockDS19,XuZHZGH018,ZhuP0019}, VAEs \citep{VahdatK20,RazaviOV19}, Flow models \citep{abs-2212-10688,BashiriWLJMDDTS21} and the current state-of-the-art diffusion models \citep{abs-2102-05379,abs-2308-05095,mou2023t2i,abs-2212-05032,RombachBLEO22}. 
Especially, the diffusion-based methods have recently delivered a remarkable performance in a plethora of cross-modal generation tasks, such as DALL-E \citep{RameshPGGVRCS21}, Stable Diffusion \citep{RombachBLEO22}.
While all previous efforts of cross-modal learning are limited to the comprehension of multimodal inputs only, CoDi \citep{abs-2305-11846} lately presents groundbreaking development.
Leveraging the power of diffusion models, CoDi possesses the ability to generate any combination of output modalities, including language, images, videos, or audio, from any combination of input modalities in parallel. 
Regrettably, CoDi might still fall short of achieving human-like deep reasoning of input content, with only parallel cross-modal feeding\&generation.

\vspace{-3mm}
\paragraph{Multimodal Large Language Models}

LLMs have already made profound impacts and revolutions on the entire AI community and beyond.
The most notable LLMs, i.e., OpenAI's ChatGPT \citep{chatgpt} and GPT4 \citep{gpt4}, with alignment techniques such as instruction tuning \citep{Ouyang0JAWMZASR22,abs-2308-10253,abs-2306-17107,abs-2304-08485} and reinforcement learning from human feedback (RLHF) \citep{StiennonO0ZLVRA20}, have demonstrated remarkable language understanding and reasoning abilities.
And a series of open-source LLMs, e.g., Flan-T5 \citep{abs-2210-11416}, Vicuna \citep{vicuna}, LLaMA \citep{abs-2302-13971} and Alpaca \citep{alpaca}, have greatly spurred advancement and made contributions to the community \citep{abs-2304-10592,abs-2305-01278}. 
Afterward, significant efforts have been made to construct LLMs dealing with multimodal inputs and tasks, leading to the development of MM-LLMs.

On the one hand, most of the researchers build fundamental MM-LLMs by aligning the well-trained encoders of various modalities to the textual feature space of LLMs, so as to let LLMs perceive other modal inputs \citep{abs-2302-14045,abs-2304-10592,abs-2205-02655,abs-2305-17216}.
For example, Flamingo \citep{AlayracDLMBHLMM22} uses a cross-attention layer to connect a frozen image encoder to the LLMs. 
BLIP-2 \citep{0008LSH23} employs a Q-Former to translate the input image queries to the LLMs.
LLaVA \citep{abs-2304-08485} employs a simple projection scheme to connect image features into the word embedding space.
There are also various similar practices for building MM-LLMs that are able to understand videos (e.g., Video-Chat \citep{abs-2305-06355} and Video-LLaMA \citep{abs-2306-02858}), audios (e.g., SpeechGPT \citep{abs-2305-11000}), etc.
Profoundly, PandaGPT \citep{abs-2305-16355} achieves a comprehensive understanding of six different modalities simultaneously by integrating the multimodal encoder, i.e., ImageBind \citep{abs-2305-05665}.

Nevertheless, these MM-LLMs all are subject to the limitation of only perceiving multimodal data, without generating content in arbitrary modalities.
To achieve LLMs with both multimodal input and output, some thus explore employing LLMs as decision-makers, and utilizing existing off-the-shelf multimodal encoders and decoders as tools to execute multimodal input and output, such as Visual-ChatGPT \citep{abs-2303-04671}, HuggingGPT \citep{abs-2303-17580}, and AudioGPT \citep{abs-2304-12995}.
As aforementioned, passing messages between modules with pure texts (i.e., LLM textual instruction) under the discrete pipeline scheme will inevitably introduce noises.
Also lacking comprehensive tuning across the whole system significantly limits the efficacy of semantics understanding.
Our work takes the mutual benefits of both the above two types, i.e., learning an any-to-any MM-LLM in an end-to-end manner.

\section{Implementation Details}
\label{sec:imp}

\subsection{Detailed Input Projection Layer}
\label{detailed:input_proj}

Through multimodal encoder, we can obtain patch-level multimodal tokens, denoting as $\bm{X}^{*} = \{\bm{x}^{*}_i\}_{i=1}^{N^{*}}$, where $* \in \{i, a, v\}$ represents image, audio, and video, respectively. 
For brevity, we eschew modal-specific notation.
Differing from the existing works that directly embed multimodal tokens into LLMs by a linear projection layer, we propose a multi-stage grouping mechanism, where patch-level tokens are grouped into concept-level tokens to facilitate the subsequent cross-modal interaction. 
Formally, we apply $L$ grouping stages, and in each stage, we randomly initialize $M_l$ learnable concept tokens $\bm{C}^{l} = \{c_j\}_{j}^{M_l}$. 
Then, we concatenate input features $\bm{X}^{l}$ and $\bm{C}^{l}$ together and then input them into a transformer layers:
$\hat{\bm{C}}^{l}, \hat{\bm{X}}^{l} = \text{Transformer}([\bm{C}^{l}; \bm{X}^{l}])$, where $\bm{X}^1 = \bm{X}$, and $[;]$ denotes the concatenation operator. 
Within $l$ grouping block, we group the updated $M_{l}$ concept tokens $\hat{\bm{X}}^{l}$ into $M_{l+1}$ new concept tokens $\hat{\bm{X}}^{l+1}$ based on the feature similarity.

Specifically, we firstly compute a similarity matrix $\bm{A}^{l}$ between $\hat{\bm{C}}^{l}$ and $\hat{\bm{X}}^{l}$ via a Gumbel-softmax: $\bm{A}^{l} = \text{Softmax}((\text{Norm}(\hat{\bm{C}}^{l}) \text{Norm}(\hat{\bm{X}}^{l}) + G)/\tau)$, where $G$ are i.i.d random samples drawn from the $\text{Gumbel}(0, 1)$ distribution, and $\tau$ is the learnable significance coefficient to assist to find a more suitable assign boundary. 
We compute the group to assign a concept token by taking the one-hot operation on the argmax over all the groups. 
Since the one-hot assignment operation via argmax is not differentiable, we instead use the straight-through trick to compute the assignment matrix as $\hat{\bm{A}}^{l} = \text{Onehot}(\text{Argmax}(\bm{A}^{l})) + \bm{A}^{l} - \text{Sg}(\bm{A}^{l})$, where $\text{Sg}(.)$ is the stop gradient operator. 
Finally, we integrate the features to updated concept tokens: 
$\bm{X}^{l+1} = \hat{\bm{C}}^{l} + \text{MLP}(\hat{\bm{A}}^{l}, \hat{\bm{X}}^{l})$. 
After $L$ stages grouping, we can obtain $M_L$ concept tokens $\bm{X}^{L}$, which are then fed into the LLM for perception and reasoning.

\subsection{Model Training}
\label{model_training}
For NExT-GPT model training, we consider a three-stage learning process:
\begin{compactitem}

    \item \textbf{Stage-1: Encoding-side Alignment Learning}. As discussed in Section $\S$\ref{enc-alignment}, we bridge the alignment to perform the caption generation task. The cross-entropy is employed as the loss function. During training, we only keep the input projection layer trainable while the other part of NExT-GPT is frozen. We employ Adam \citep{KingmaB14} optimizer to update the parameters. This stage can be understood as training a compatible multimodal tokenizer for the frozen LLM. 
    
    \item \textbf{Stage-2: Decoding-side Alignment Learning}. The output projection layer adopts a transformer-based architecture characterized by a hidden size of 512, 4 attention heads, 4 encoder layers, and 4 decoder layers. Additionally, the dropout ratio is set as 0.1.
    The optimization process for the three output projection layers involves a combination of three training objectives: cross-entropy focusing on the generated signal tokens, $\textit{l}_2$-distance measuring the alignment between the representation of signal tokens and captions, and conditional latent denoising loss, as shown in Section $\S$\ref{dec-alignment}. We employ the Adam optimizer for this stage, with only the parameters of the output projection layers being learnable, while others remain frozen. 
    
    \item \textbf{Stage-3: End-to-end Instruction-Tuning}. In this stage. we train the whole NExT-GPT using instruction-following datasets, as enumerated in Section $\S$\ref{Instruction Dataset}.  We incorporate LoRA to fine-tune the weights of the LLM. Moreover, both the input and output projection layers are trainable. The training objectives include two parts: 1) cross-entropy between the generated and gold response, 2) generation loss. The Adam optimizer is applied to update the learnable parameters.
    
\end{compactitem}

\subsection{Detailed Dataset}
\label{detailed_dataset}
Here, we enumerate the datasets employed for training and fine-tuning NExT-GPT:
\begin{compactitem}
    \item \textbf{`Text-X' Pair Dataset}. 
    \begin{compactitem}
        \item \textbf{CC3M} \citep{SoricutDSG18}: contains over 3 million images accompanied by diverse styles of natural-language descriptions.
        \item \textbf{COCO-caption} \citep{LinMBHPRDZ14}: is a large-scale image-text pair dataset which is taken as image captioning, or text-to-image generation task benchmark.
        \item \textbf{WebVid-2M} \citep{BainNVZ21}: is a large-scale dataset of short videos with textual description sourced from stock footage sites.
        \item \textbf{AudioCaps} \citep{KimKLK19}: with 46K audio-text pairs derived from the AudioSet \citep{GemmekeEFJLMPR17} dataset.
    \end{compactitem}
    
    \item \textbf{`Text' $\to$ `Text' Instruction Datasets}.
     \begin{compactitem}
    \item  \textbf{Cleaned-Alpaca}\footnote{\url{https://github.com/gururise/AlpacaDataCleaned}}: is a textual instruction dataset used to train the Alpaca LLM (Large Language Model).
     \end{compactitem}
      
      \item \textbf{`Text+X' $\to$ `Text' Instruction Datasets}.
     \begin{compactitem} 
     \item \textbf{LLaVA-150K} \citep{abs-2304-08485}: is a set of GPT-generated multimodal instruction-following data. It is constructed for visual instruction tuning and for building large multimodal towards GPT-4 vision/language capability.
     \item \textbf{VideoChat} \citep{abs-2305-06355}: comprising thousands of videos paired with detailed dataset textual descriptions and conversations generated using dense captions fed to ChatGPT in temporal order.   
     \end{compactitem}

    \item \textbf{Modality-switching Dataset}
    \begin{compactitem}
        \item \textbf{MosIT}: encompasses a wide range topic of dialogues between `Human' and `Machine'. Each dialogue includes 3-7 turns (i.e., QA pairs), where the `Human'-`Machine' interactions should involve multiple modalities at either the input or output side, and switch the modalities alternately. 
    \end{compactitem}
    
\end{compactitem}

\subsection{Multimodal IT Datasets Comparison}
\label{Multimodal IT Datasets Comparison}

Here, we compare the existing multimodal instruction tuning (IT) datasets, as detailed in Table \ref{tab:it-data}. 
As can be seen, the response modality of the existing IT datasets is merely limited to text.
In this work, we leverage GPT-4 to generate a T2M IT dataset, comprising 15k instances, which serves as a foundation for instructing the model to generate responses in other modalities, such as image, video, and audio. 
Furthermore, we construct a modality-switching IT dataset with 5k instances, named \texttt{MosIT}. 
This dataset is designed to emulate the human-machine complex interaction featuring diverse and dynamic shifts in modalities within both inputs and outputs.

\begin{sidewaystable}[htbp]
\centering
\fontsize{9}{12.5}\selectfont
\setlength{\tabcolsep}{0.5mm}
\begin{tabular}{lccccccc}
\hline
\bf Dataset & \bf Data Source & \bf In$\to$Out Modality & \bf Approach & \bf Multi-turn Reason &	 \bf \#Img/Vid/Aud & \bf \#Dialog Turn. &	\bf \#Instance \\ \hline
\multicolumn{6}{l}{\textbf{\em $\blacktriangleright$ Existing data}} \\
MiniGPT-4 \citep{abs-2304-10592}  &		CC, CC3M  &		T+I$\to$T &		Auto	& \textcolor{ired}{\ding{55}} & 134M/-/-	 &	1 & 5K  \\
StableLLaVA \citep{abs-2308-10253} &		SD  &		T+I$\to$T	 &	Auto+Manu.	& \textcolor{ired}{\ding{55}} &	126K/-/-	 &		1	& 126K \\
LLaVA \citep{abs-2306-17107} &		COCO  &		 T+I$\to$T & Auto & \textcolor{igreen}{\ding{51}}	& 81K/-/-	 	 &	2.29	 &		150K  \\
SVIT \citep{abs-2307-04087} &		MS-COCO, VG  &		T+I$\to$T & Auto & \textcolor{igreen}{\ding{51}}	 &	108K/-/-	 &	5 & 3.2M  \\
LLaVAR \citep{abs-2306-17107} &		COCO,  CC3M , LAION &		T+I$\to$T	 &	LLaVA+Auto	& \textcolor{igreen}{\ding{51}}	 &		20K/-/-  &		2.27 & 174K \\
\cdashline{1-8}
VideoChat \citep{abs-2305-06355} &		WebVid  &		T+V$\to$T  &	Auto	 & \textcolor{igreen}{\ding{51}} &	 -/8K/-	 &	1.82 &		11K  \\
Video-ChatGPT \citep{abs-2306-05424} &		ActivityNet  &		T+V$\to$T	 & Inherit	& \textcolor{ired}{\ding{55}} & -/100K/-	 &	1  &		100K \\
Video-LLaMA \citep{abs-2306-02858} &		MiniGPT-4, LLaVA, VideoChat &		T+I/V$\to$T	 & Auto & \textcolor{igreen}{\ding{51}} &	81K/8K/-	 &	2.22	  &		 171K  \\
InstructBLIP \citep{abs-2305-06500} &	Multiple &		T+I/V$\to$T	 &	Auto	& \textcolor{ired}{\ding{55}} & -	&		- & $\sim$ 1.6M \\
MIMIC-IT \citep{abs-2306-05425} &		Multiple &		T+I/V$\to$T	 &	Auto & \textcolor{ired}{\ding{55}} &		8.1M/502K/-	 &	1	& 2.8M  \\
PandaGPT \citep{abs-2305-16355} &		MiniGPT-4, LLaVA &	T+I$\to$T  &	Inherit	& \textcolor{igreen}{\ding{51}} &	 81K/-/-	 &		2.29 &		160K  \\
\cdashline{1-8}
MGVLID \citep{abs-2307-09474} &		Multiple&		T+I+B$\to$T	 &	Auto+Manu.& \textcolor{ired}{\ding{55}}	 &	108K/-/- &	-	 & 108K \\
M$^3$IT \citep{abs-2306-04387} &		Multiple &		T+I/V/B$\to$T	 & Auto+Manu.& \textcolor{ired}{\ding{55}} &	-/-/-	 &	1  &		2.4M \\
LAMM \citep{abs-2306-06687} &		Multiple &		T+I+PC$\to$T	 &	Auto+Manu.& \textcolor{igreen}{\ding{51}}	 &		91K/-/- &	3.27	& 196k  \\
BuboGPT \citep{abs-2307-08581} &		Clotho , VGGSS  &		T+A/(I+A)$\to$T	 &	Auto & \textcolor{ired}{\ding{55}}	 &	5k/-/9K &  - &		9K \\
mPLUG-DocOwl \citep{abs-2307-02499} &		Multiple &		T+I/Tab/Web$\to$T	 &	Inherit & \textcolor{ired}{\ding{55}} & -	 &		- &		- \\

\hline\hline
\multicolumn{6}{l}{\em \textbf{$\blacktriangleright$ In this work}} \\
T2M &		Webvid , CC3M , AudioCap &		T$\to$T+I/A/V& Auto   & \textcolor{ired}{\ding{55}}&		5K/5K/5K &	1	& 15K   \\
\texttt{MosIT} &	Youtube, Google, Flickr, Midjourney, etc.  &		\cellcolor{ired} T+I+A+V$\to$T+I+A+V & Auto+Manu.& \textcolor{igreen}{\ding{51}} &		4K/4K/4K &	4.8 &		5K  \\

\hline
\end{tabular}
\caption{
Summary and comparison of existing datasets for multimodal instruction tuning.
T: text, I: image, V: video, A: audio,
B: bounding box, PC: point cloud, Tab: table, Web: web page.
}
\label{tab:it-data}
\end{sidewaystable}

\subsection{Training Recipes}

In Table \ref{tab:training_recipes}, we list the detailed hyper-parameters setting at three stages.

\begin{table*}[!th]
\centering
\fontsize{8}{11}\selectfont
\setlength{\tabcolsep}{5mm}
\caption{
\label{tab:training_recipes}
Training recipes for NExT-GPT. The three training stages are introduced in Section \ref{model_training}.
Stage-1: Encoding-side Alignment Learning, Stage-2: Decoding-side Alignment Learning, Stage-3: End-to-end Instruction Tuning.
}
\begin{tabular}{lccc}
\hline
\bf Configuration & \bf Stage-1 & \bf Stage-2 & \bf Stage-3 \\ 
\toprule
Optimizer & Adam & Adam & Adam  \\
Learning Rate & 0.0004 & 0.0004 & 0.0005\\
Weight Decay & 0.001 & 0.001 & 0.001\\
Training Epochs & 1 & 1 & 1\\
Warmup Ratio & 0.1 & 0.1 & 0.1\\
Learning Rate Scheduler & Linear & Linear & Linear \\
Batch Size Per GPU & 18 & 8 & 4\\
Maximum Token Length & 512 & 512 & 512\\
Unfreeze LLM & \usym{2717}  & \usym{2717} & \usym{2714} \\
\hline
\multicolumn{4}{c}{\bf Training Data} \\
\hline
\multirow{3}{*}{Dataset} & {CC3M} & {CC3M} & LLaVA-150K, VideoChat  \\
& WebVid & WebVid  & cleaned-Alpaca \\
& AudioCaps & AudioCaps & Text$\rightarrow$Text+X, MosIT  \\
\hline
\end{tabular}
\vspace{-1mm}
\end{table*}

\begin{figure}[!h]
\centering
\includegraphics[width=1.0\columnwidth]{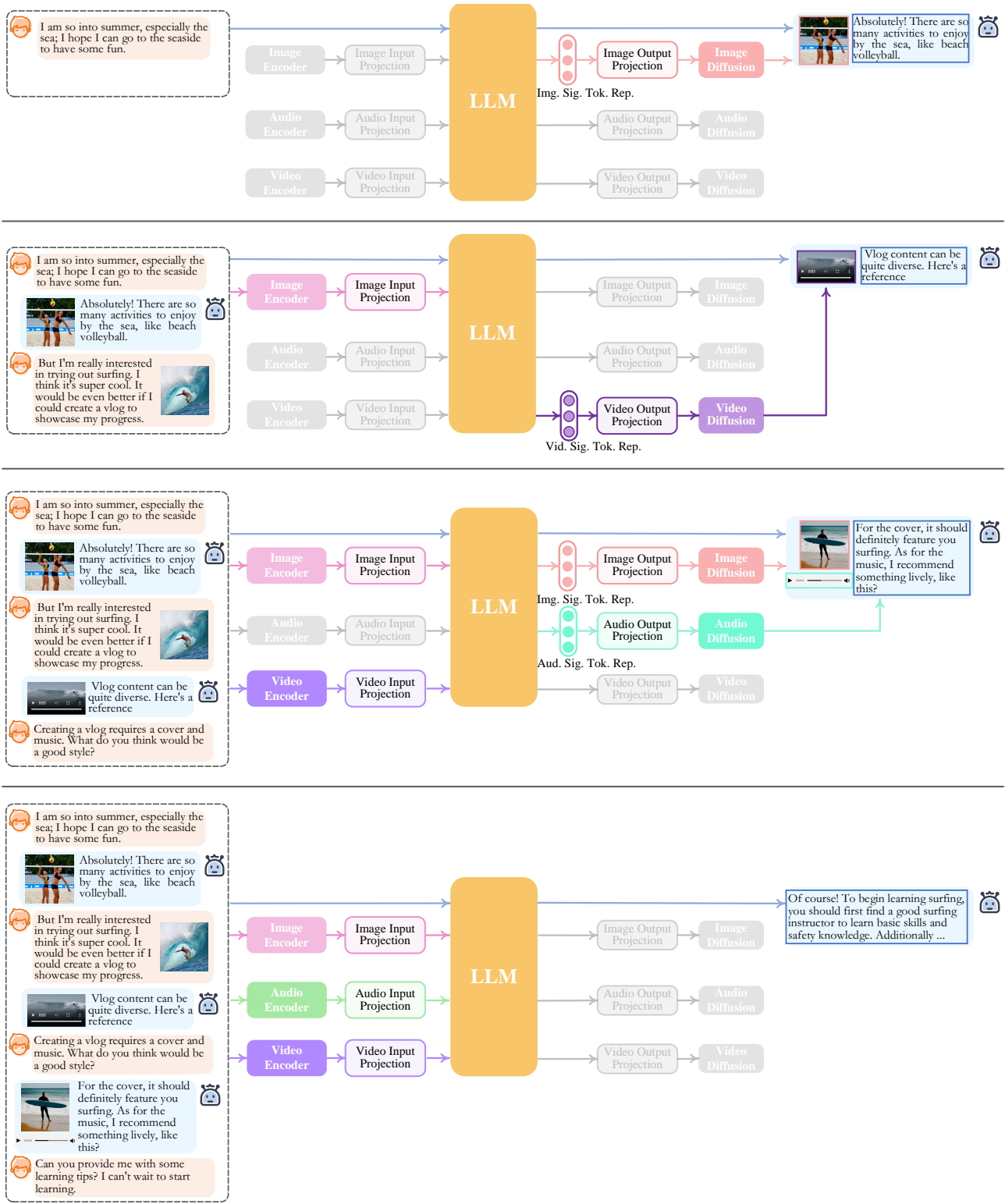}
\caption{
NExT-GPT inference process.
Grey colors denote the deactivation of the modules.
}
\label{fig:inference}
  \vspace{-3mm}
\end{figure}

\vspace{-2mm}
\subsection{Inference Process}

In Figure \ref{fig:inference} we further illustrate the inference procedure of NExT-GPT.
Given certain user inputs of any combination of modalities, the corresponding modal encoders, and projectors transform them into feature representations and pass them to LLM\footnote{Except the text inputs, which will be directly fed into LLM.}.
Then, LLM decides what content to generate, i.e., textual tokens, and modality signal tokens.
If LLM identifies a certain modality content (except language) to be produced, a special type of token \citep{abs-2305-17216} will be output indicating the activation of that modality; otherwise, no special token output means deactivation of that modality.

\section{Additional Experiments}
\label{Additional Experiments}

 \subsection{Additional Multimodal Comprehension and Generation Results}
Firstly, we examine the synthesis quality of the image, video, or audio conditioned on text compared with the non-LLM-based methods. 
Table \ref{tab:T2I-res}, \ref{tab:T2A-res}, \ref{tab:T2V-res} present the comparisons between ours and some state-of-the-art systems.
On text-to-image and text-to-audio generation tasks, NExT-GPT shows a nice performance on par with that of the best-performing baselines.
Notably, under the zero-shot setting, NExT-GPT shows a significant superiority in video generation conditioning on text, demonstrating the remarkable generalization capability of NExT-GPT.

Secondly, we evaluate the NExT-GPT on the tasks of textual caption generation to test the semantic understanding capability w.r.t. image, video, or audio.
The results on different tasks are shown in Table \ref{tab:I2T-res}, \ref{tab:A2T-res}, and \ref{tab:V2T-res}.
Significantly, NExT-GPT mostly achieves much better performance on the X-to-text generation than that of the CoDi baseline, owing to the direct generation of texts from LLM, which is inherently expertized by the LLM.

Thirdly, we test our model on text-conditioned modal editing tasks.
Table \ref{tab:TI2I-res}, \ref{tab:TA2A-res} and \ref{tab:TV2V-res} show the performances on different tasks.
Compared with the above two types of tasks, although NExT-GPT did not demonstrate superior performance on the text-conditioned modal editing tasks, it still shows competitive performance.

\begin{table*}[!t]
\begin{minipage}{\textwidth}
\centering
\begin{minipage}[b]{0.36\textwidth}
\fontsize{8}{11}\selectfont
\setlength{\tabcolsep}{3.5mm}
\centering
\caption{
\label{tab:T2I-res}
Text-to-image generation results on COCO-caption \citep{LinMBHPRDZ14}.
}
\begin{tabular}{lc}
\hline
\bf Method & \bf FID ($\downarrow$) \\
\hline
CogView \citep{DingYHZZYLZSYT21}& 	27.10 \\   
GLIDE \citep{NicholDRSMMSC22} & 	12.24 \\
CoDi \citep{abs-2305-11846} &  11.26 \\
SD \citep{RombachBLEO22} &  11.21 \\
\cdashline{1-2}
\rowcolor{bluei} NExT-GPT & \bf 11.18\\
\hline
\end{tabular}
\end{minipage}
\hfill
\begin{minipage}[b]{0.45\textwidth}
\centering
\fontsize{8}{11}\selectfont
\setlength{\tabcolsep}{1.1mm}
\caption{
\label{tab:T2V-res}
Text-to-video generation results (zero-shot) on MSR-VTT \citep{XuMYR16}.
}
\begin{tabular}{lcc}
\hline
\bf Method & \bf FID ($\downarrow$) & \bf CLIPSIM ($\uparrow$) \\
\hline
CogVideo \citep{abs-2205-15868}&	23.59 & 	26.31 \\   
MakeVideo \citep{abs-2209-14792} & 	13.17 & 	30.49 \\   
Latent-VDM \citep{RombachBLEO22} &  14.25 & 	27.56 \\   
Latent-Shift \citep{abs-2304-08477} & 15.23 & 	27.73 \\
CoDi \citep{abs-2305-11846} & --- &  28.90 \\
\cdashline{1-3}
\rowcolor{bluei} NExT-GPT & \bf 12.69 &  \bf 31.97 \\    
\hline
\end{tabular}
\end{minipage}
\end{minipage}
\vspace{-2mm}
\end{table*}

\begin{table*}[!t]
\begin{minipage}{\textwidth}
\centering
\begin{minipage}[b]{0.36\textwidth}
\centering
\fontsize{8}{11}\selectfont
\setlength{\tabcolsep}{1.5mm}
\caption{
\label{tab:T2A-res}
Text-to-audio generation results on AudioCaps \citep{KimKLK19}.
}
\begin{tabular}{lcc}
\hline
\bf Method & \bf FD ($\downarrow$) & \bf IS ($\uparrow$) \\
\hline
DiffSound \citep{YangYWWWZY23}&	47.68 & 4.01 \\   
AudioLDM-S \citep{LiuCYMLM0P23} & 29.48 & 6.90 \\   
AudioLDM-L \citep{LiuCYMLM0P23} &  23.31  &  8.13 \\   
CoDi \citep{abs-2305-11846} & \bf 22.90 & \bf 8.77 \\
\cdashline{1-3}
\rowcolor{bluei} NExT-GPT & 23.25 & 8.67 \\    
\hline
\end{tabular}
\end{minipage}
\hfill
\begin{minipage}[b]{0.45\textwidth}
\centering
\fontsize{8}{11}\selectfont
\setlength{\tabcolsep}{1.5mm}
\caption{
\label{tab:A2T-res}
Audio-to-text generation (audio captioning) results on AudioCaps \citep{KimKLK19}.
}
\begin{tabular}{lcc}
\hline
\bf Method & \bf SPIDEr & \bf CIDEr \\
\hline
AudioCaps \citep{KimKLK19}&	 0.369& 0.593 \\   
BART \citep{GontierSC21} &  0.465& 0.753 \\   
AL-MixGen \citep{abs-2210-17143} &  0.466 & 0.755 \\   
CoDi \citep{abs-2305-11846} & 0.480 & 0.789 \\
\cdashline{1-3}
\rowcolor{bluei} NExT-GPT &  \bf  0.534 & \bf  0.807 \\    
\hline
\end{tabular}
\end{minipage}
\end{minipage}
\end{table*}

\begin{table*}[!t]
\begin{minipage}{\textwidth}
\centering
\begin{minipage}[b]{0.50\textwidth}
\fontsize{8}{11}\selectfont
\setlength{\tabcolsep}{1.2mm}
\centering
\caption{
\label{tab:I2T-res}
Image-to-text generation (image captioning) results on COCO-caption \citep{LinMBHPRDZ14}.
}
\begin{tabular}{lccc}
\hline
\bf Method & \bf B@4 & \bf METEOR & \bf CIDEr \\
\hline
Oscar \citep{Li0LZHZWH0WCG20}  &  36.58  & 30.4  & 124.12 \\
BLIP-2 \citep{0008LSH23} &  43.7 &  --- &  145.8 \\
OFA \citep{WangYMLBLMZZY22} &   \bf44.9 &  32.5 &  154.9 \\
CoDi \citep{abs-2305-11846} &   40.2 &  31.0 &  149.9 \\
\cdashline{1-4}
\rowcolor{bluei} NExT-GPT & 45.1   & \bf 34.1  & \bf 158.3  \\
\hline
\end{tabular} 
\end{minipage}
\hfill
\begin{minipage}[b]{0.47\textwidth}
\centering
\fontsize{8}{11}\selectfont
\setlength{\tabcolsep}{1.1mm}
\caption{
\label{tab:V2T-res}
Video-to-text generation (video captioning) results on MSR-VTT \citep{XuMYR16}.
}
\begin{tabular}{lcc}
\hline
\bf Method & \bf B@4 & \bf METEOR  \\
\hline
ORG-TRL \citep{ZhangSY0WHZ20}&	43.6 &  28.8 \\   
GIT \citep{WangYHLLGLLW22} & 	 54.8 &  33.1 \\   
mPLUG-2 \citep{XuYYSYXLBQWXZH023} &  57.8 &  34.9 \\   
CoDi \citep{abs-2305-11846} & 52.1 &  32.5 \\
\cdashline{1-3}
\rowcolor{bluei} NExT-GPT & \bf 58.8 &  \bf 39.6 \\    
\hline
\end{tabular}
\end{minipage}
\vspace{-2mm}
\end{minipage}
\end{table*}

\begin{table*}[!t]
\begin{minipage}{\textwidth}
\centering
\begin{minipage}[b]{0.51\textwidth}
\fontsize{8}{11}\selectfont
\setlength{\tabcolsep}{0.4mm}
\centering
\caption{
\label{tab:TI2I-res}
Text+image-to-image generation (text-conditioned image editing) results on COCO-caption \citep{LinMBHPRDZ14}.
}
\begin{tabular}{lcccc}
\hline
\multirow{2}{*}{\bf Method}
& \multicolumn{2}{c}{\bf Object} & \multicolumn{2}{c}{\bf Background} \\ 
\cmidrule(r){2-3} \cmidrule(r){4-5}
 & \bf CLIP ($\uparrow$) & \bf FID ($\downarrow$)  & \bf CLIP ($\uparrow$) & \bf FID ($\downarrow$) \\
\hline
PTP \citep{HertzMTAPC23}  &  30.33 &  9.58  & 31.55 &  13.92 \\
BLDM \citep{AvrahamiFL23} &  29.95 &  6.14 &  30.38 &  20.44 \\
DiffEdit \citep{CouaironVSC23} &   29.30 &\bf 3.78 &  26.92 &  \bf 1.74 \\
PFB-Diff \citep{abs-2306-16894} &\bf  30.81 &  5.93 &  \bf 32.25 &  13.77 \\
\cdashline{1-5}
\rowcolor{bluei} NExT-GPT & 29.32   & 6.62 & 27.31  & 14.27  \\
\hline
\end{tabular}
\end{minipage}
\hfill
\begin{minipage}[b]{0.46\textwidth}
\centering
\fontsize{8}{11}\selectfont
\setlength{\tabcolsep}{0.3mm}
\caption{
\label{tab:TV2V-res}
Text+video-to-video generation (text-conditioned video editing) results on DAVIS \citep{PerazziPMGGS16}.
}
\begin{tabular}{lcc}
\hline
\bf Method & \bf CLIP-T & \bf CLIP-I  \\
\hline
CogVideo \citep{abs-2205-15868}&	0.2391 &   0.9064  \\   
TuneVideo \citep{abs-2212-11565} & 	  0.2758 & 0.9240 \\   
SDEdit \citep{MengHSSWZE22} &  0.2775 & 0.8731 \\   
Pix2Video \citep{abs-2303-12688} & \bf 0.2891  & \bf 0.9767 \\
\cdashline{1-3}
\rowcolor{bluei} NExT-GPT & 0.2684 &  0.9647 \\    
\hline
\end{tabular}
\end{minipage}
\end{minipage}
\vspace{-1mm}
\end{table*}

\begin{table}[!h]
\centering
\fontsize{8}{11}\selectfont
\setlength{\tabcolsep}{8mm}
\caption{
\label{tab:TA2A-res}
Text+audio-to-audio generation (text-conditioned speech editing) results on VCTK \citep{veaux2017cstr}.
}
\begin{tabular}{lc}
\hline
\bf Method & \bf MCD ($\downarrow$) \\
\hline
CampNet \citep{WangYFTW22}& 0.380 \\   
MakeAudio \citep{HuangHY0LLYLYZ23}  & 0.375 \\   
AudioLDM-L \citep{LiuCYMLM0P23}  & 0.349 \\
\cdashline{1-2}
\rowcolor{bluei} NExT-GPT &  \bf 0.300 \\    
\hline
\end{tabular}
\vspace{-4mm}
\end{table}

\subsection{Human Evaluation on Complex Any-to-any QA}
\begin{figure}[!h]
\centering
\includegraphics[width=0.75\columnwidth]{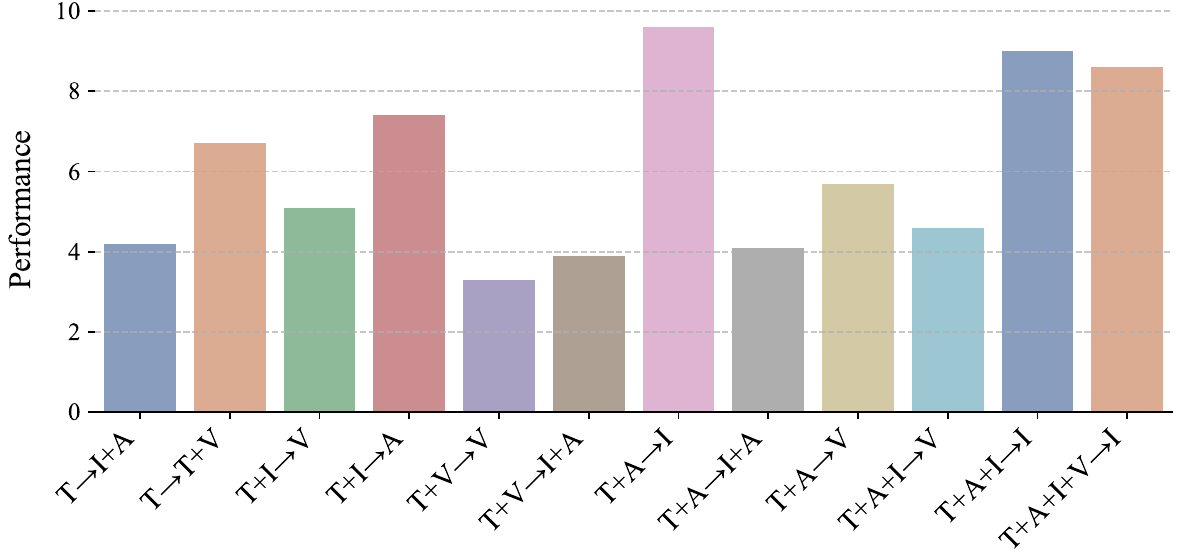}
\vspace{-4mm}
\caption{
Comparative performance of NExT-GPT on various complex cross-modal conversions.
}
\label{fig:human-eval}
\vspace{-3mm}
\end{figure}

We also carry out evaluation on some more scenarios where there are complicated cross-modal interactions between inputs and outputs.
We mainly compare the model performance for the settings with different modality conversions.
As no standard benchmark can be leveraged, here we adopt human evaluation.
We ask several evaluators to score the performance of NExT-GPT on a scale from 1 to 10.
Figure \ref{fig:human-eval} shows the comparisons.
We find NExT-GPT is more competent in producing images, compared with the generations on videos and audio.
Also generating mixed combinations of multimodal content is slightly inferior to the generation of single-modal content, due to the complexity of the latter.

\subsection{Case Study on Pipeline-style vs. End-to-end Unification}
We earlier have elaborated on the difference as well as the necessity of building a unified any-to-any multimodal LLM in an end-to-end manner, compared with the existing pipeline-style systems that generate intermediate captions and then pass to the downstream tools (e.g., diffusion models for generation).
The cascade process inevitably introduces noise and propagates errors.
Meanwhile, the entire system only leverages existing pre-trained tools for inference, whereas, without end-to-end updating throughout the whole system, the capability to more accurately interpret complex user instructions and generate content will be compromised.
Here we add a few illustrations, where we make comparisons with these pipeline-style systems:
1) Visual-ChatGPT and HuggingGPT, which are existing systems that have free open access; 
2) NExT-GPT variant with captions as the messenger (which we mark as NExT-GPT-caption). 
To implement NExT-GPT-caption, the captions directly generated by LLM will be fed into the following generation models, instead of using the soft representations of the signal tokens. 
As Visual-ChatGPT only supports image generation, we here consider the evaluation on the Text-to-Text\&Image setting.

\begin{figure}[!h]
\centering
\includegraphics[width=0.90\columnwidth]{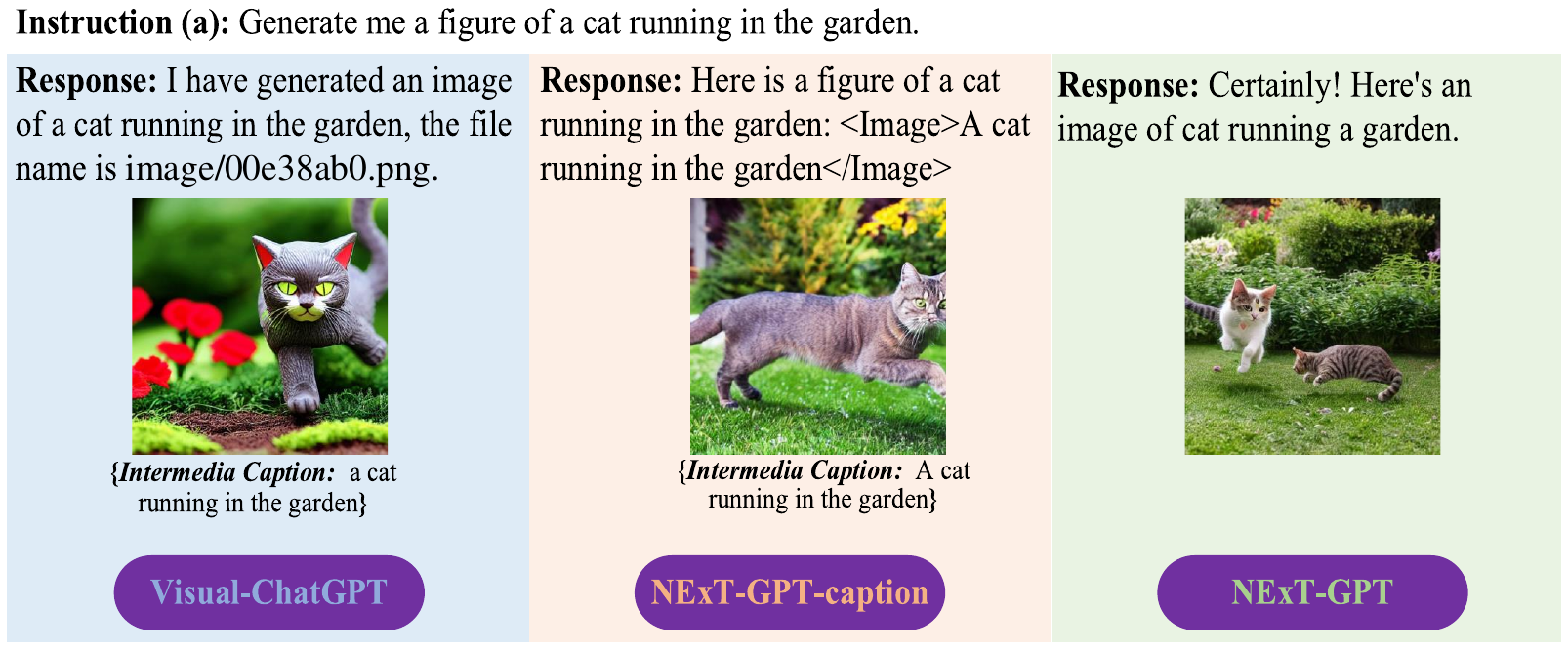}
\caption{
Illustration of case study, image generation from a simple instruction on Visual-ChatGPT, NExT-GPT-caption, and NExT-GPT. 
}
\label{fig:pipeline-end2end-1}
\end{figure}

\begin{figure}[!h]
\centering
\includegraphics[width=0.89\columnwidth]{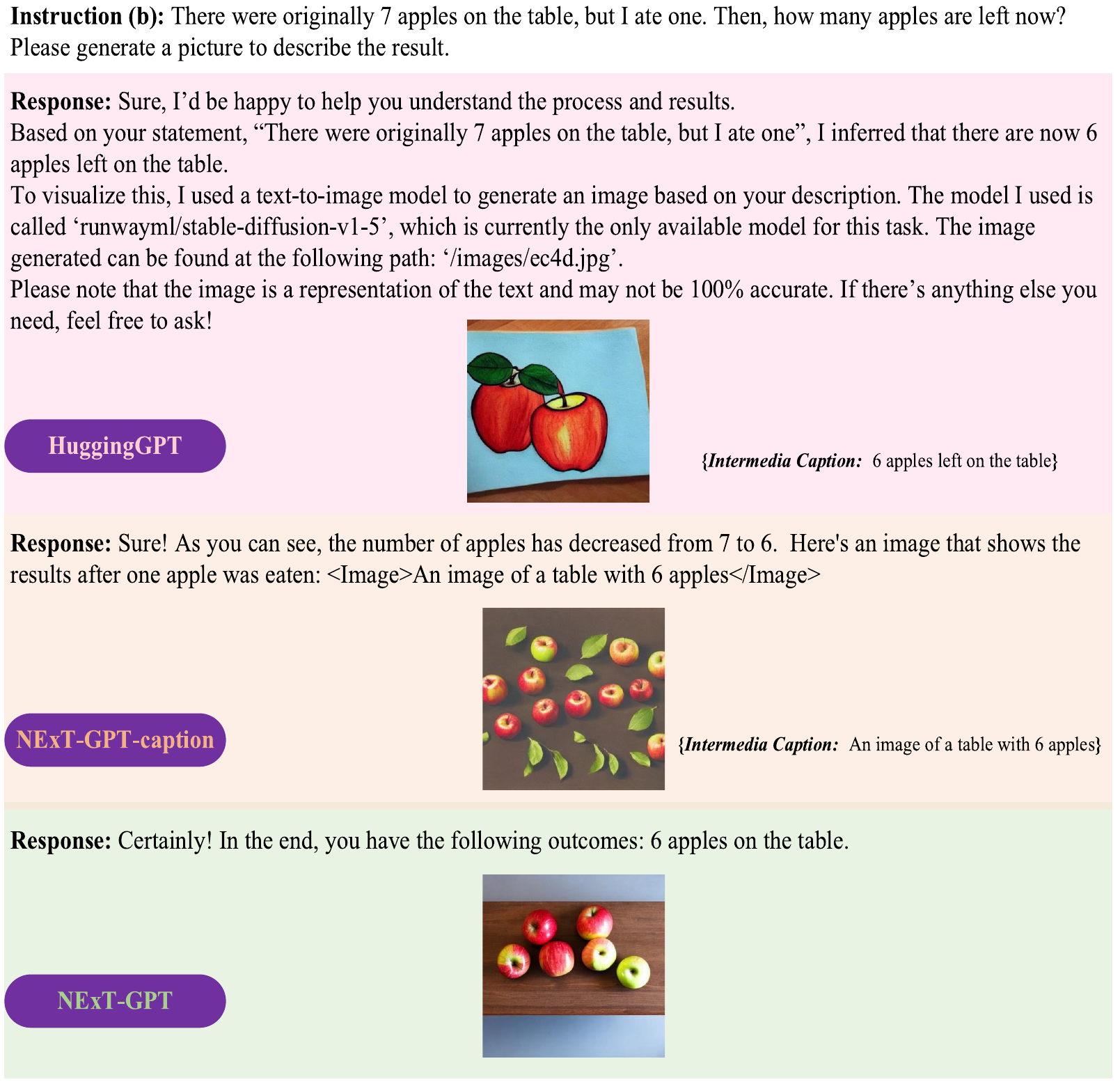}
\vspace{-2mm}
\caption{
Illustration of image generation from a complex instruction on HuggingGPT, NExT-GPT-caption, and NExT-GPT. 
In this case, the numeration understanding of vision has been wrong by pipeline-style methods.
}
\label{fig:pipeline-end2end-2}
\end{figure}

\begin{figure}[!t]
\centering
\includegraphics[width=0.89\columnwidth]{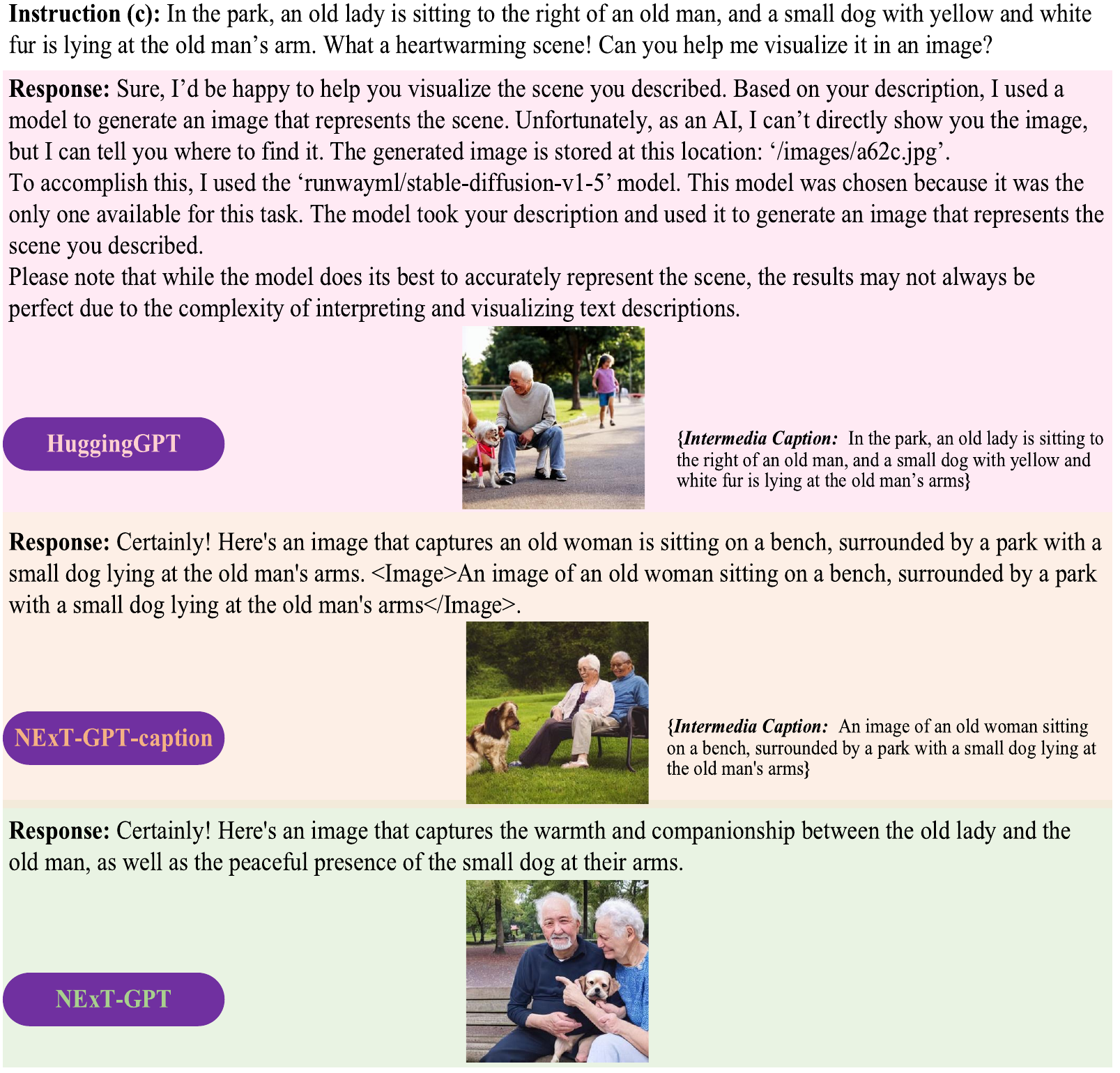}
\vspace{-2mm}
\caption{
Illustration of image generation from another complex instruction on HuggingGPT, NExT-GPT-caption, and NExT-GPT.
In this case, the understanding of visual-spatial relational semantics has been wrong by pipeline-style methods.
}
\label{fig:pipeline-end2end-3}
\end{figure}

Figure \ref{fig:pipeline-end2end-1} presents the case of image generation from a simple input user instruction; while Figure \ref{fig:pipeline-end2end-2} and \ref{fig:pipeline-end2end-3} present two cases of image generation from comparatively complex input user instructions.
On the simple one, all generated image content from both pipeline-style and end-to-end (ours) systems seem correct and coincide with the input prompt.
However, when handling complex instructions, as seen in Figure \ref{fig:pipeline-end2end-2} and \ref{fig:pipeline-end2end-3}, the generated image content can be wrong and biased to the user's intention.
The problems are rooted in the core of different modalities, i.e., there are inherent gaps between language and visual modalities that cannot be eliminated. 
Here are two representative attributes: \textbf{the numeration of vision} (cf. Figure \ref{fig:pipeline-end2end-2}) and \textbf{the visual-spatial relational semantics} (cf. Figure \ref{fig:pipeline-end2end-3}), which could be hard to (or even cannot) be expressed by the intermediate captions conveniently.
Utilizing textual captions as intermediate representations runs the risk of overlooking these modality-specific features when expressing non-linguistic (e.g., visual) modalities solely through language.

By the way, we kindly note a fact that, with the intermediate captions produced from the pipeline-style systems in Figure \ref{fig:pipeline-end2end-2} and \ref{fig:pipeline-end2end-3}, the Stable Diffusion model just has difficulty in accurately understanding the vision numeration and visual-spatial relation and generating correct answers, i.e., they are the problems 
inherent to the Stable Diffusion model itself, and Stable Diffusion alone is tricky to overcome.
Most recent work tries to solve this issue by integrating the vision-specific features into the Stable Diffusion \citep{abs-2305-15393,QuW0NC23} via additional feature engineering.
But, in our NExT-GPT with an end-to-end solution, the implicit modality signal token embeddings that carry rich modality-specific features of non-linguistic will be naturally encoded and passed to the downstream modules (e.g., Stable Diffusion), without any further external effort.

\vspace{-2mm}
\subsection{Example Demonstrations}
\label{Example Demonstrations}

\begin{figure}[!ht]
\centering
\includegraphics[width=1\columnwidth]{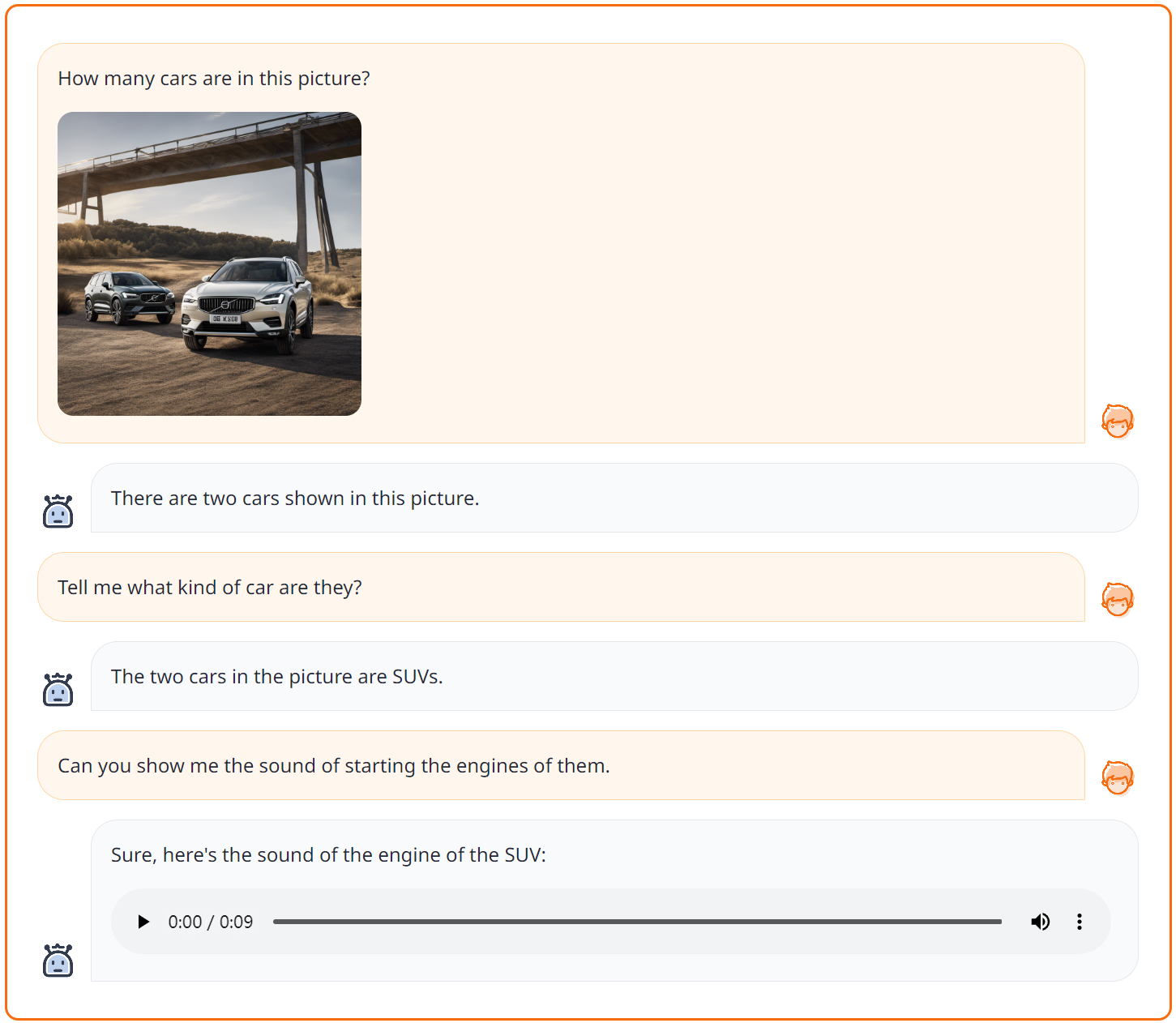}
\caption{
Example of Text+Image $\to$ Text+Audio.
}
\label{fig:T+I-T+A}
\end{figure}

\begin{figure}[!ht]
\centering
\includegraphics[width=0.89\columnwidth]{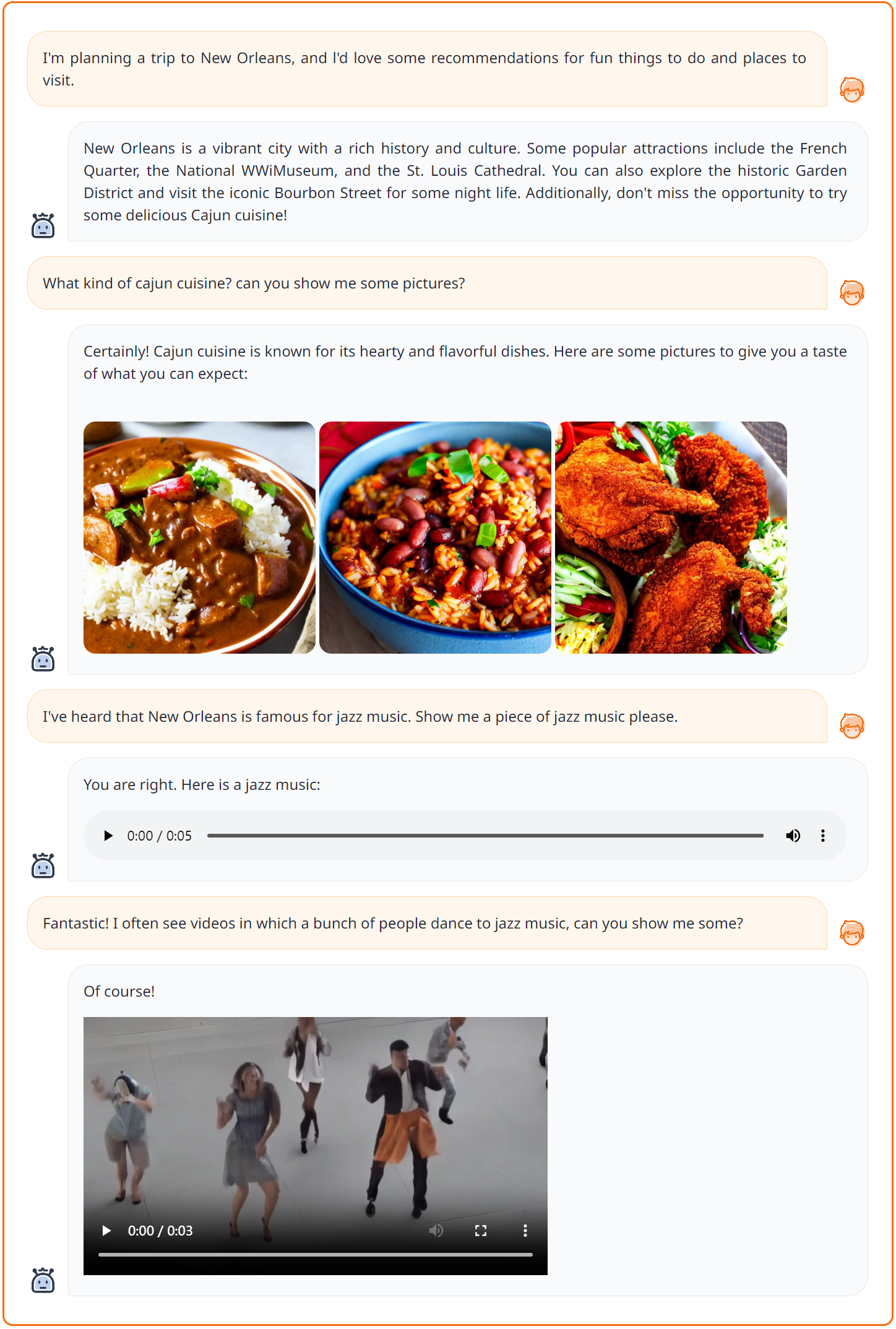}
\caption{
Example of Text $\to$ Text+Image+Video+Audio.
}
\label{fig:T-T+I+V+A}
\end{figure}

\begin{figure}[!ht]
\vspace{-2mm}
\centering
\includegraphics[width=0.80\columnwidth]{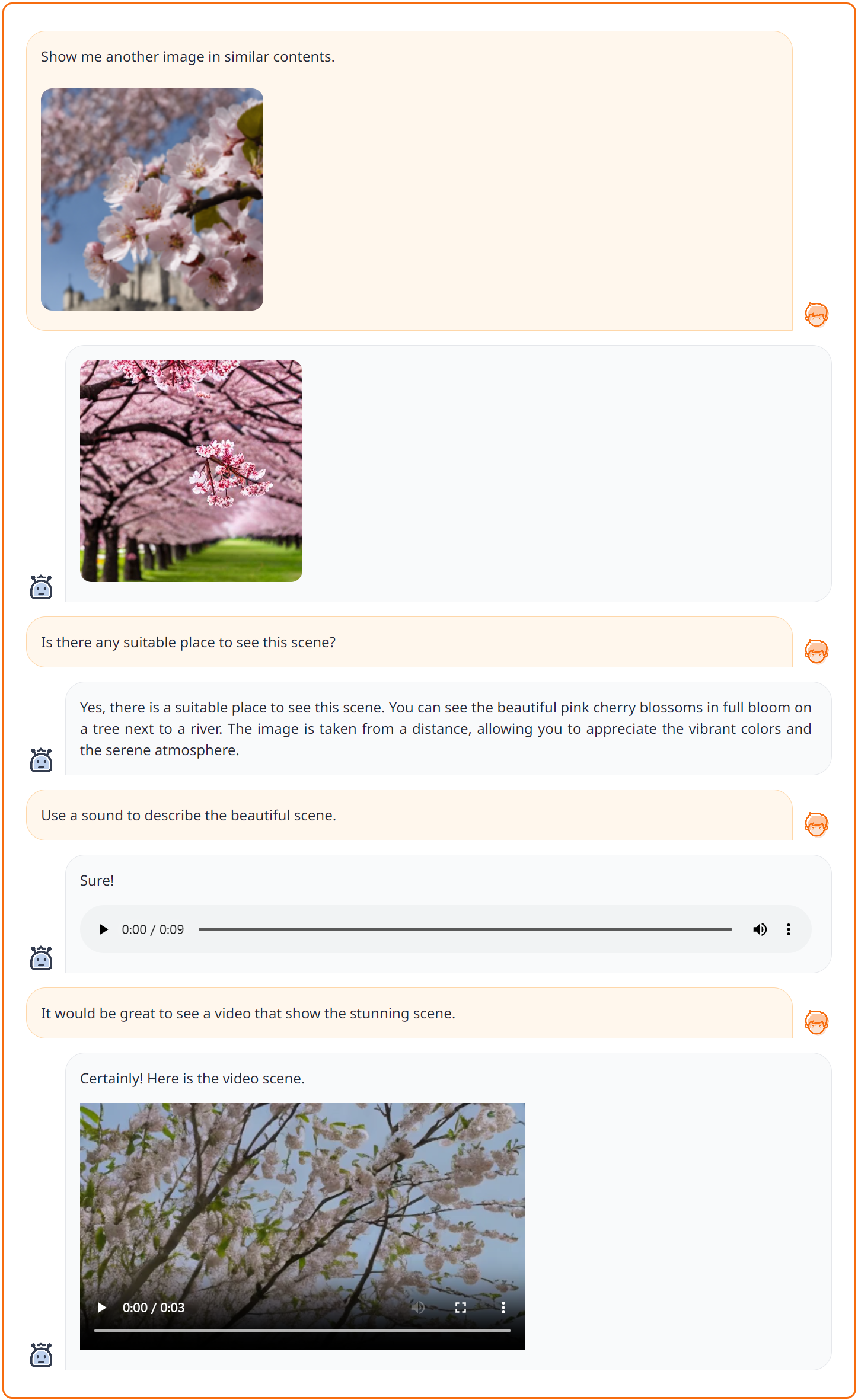}
\caption{
Example of Text+Image $\to$ Text+Image+Video+Audio.
}
\label{fig:T+I-T+I+V+A}
\vspace{-3mm}
\end{figure}

\begin{figure}[!ht]
\centering
\includegraphics[width=0.92\columnwidth]{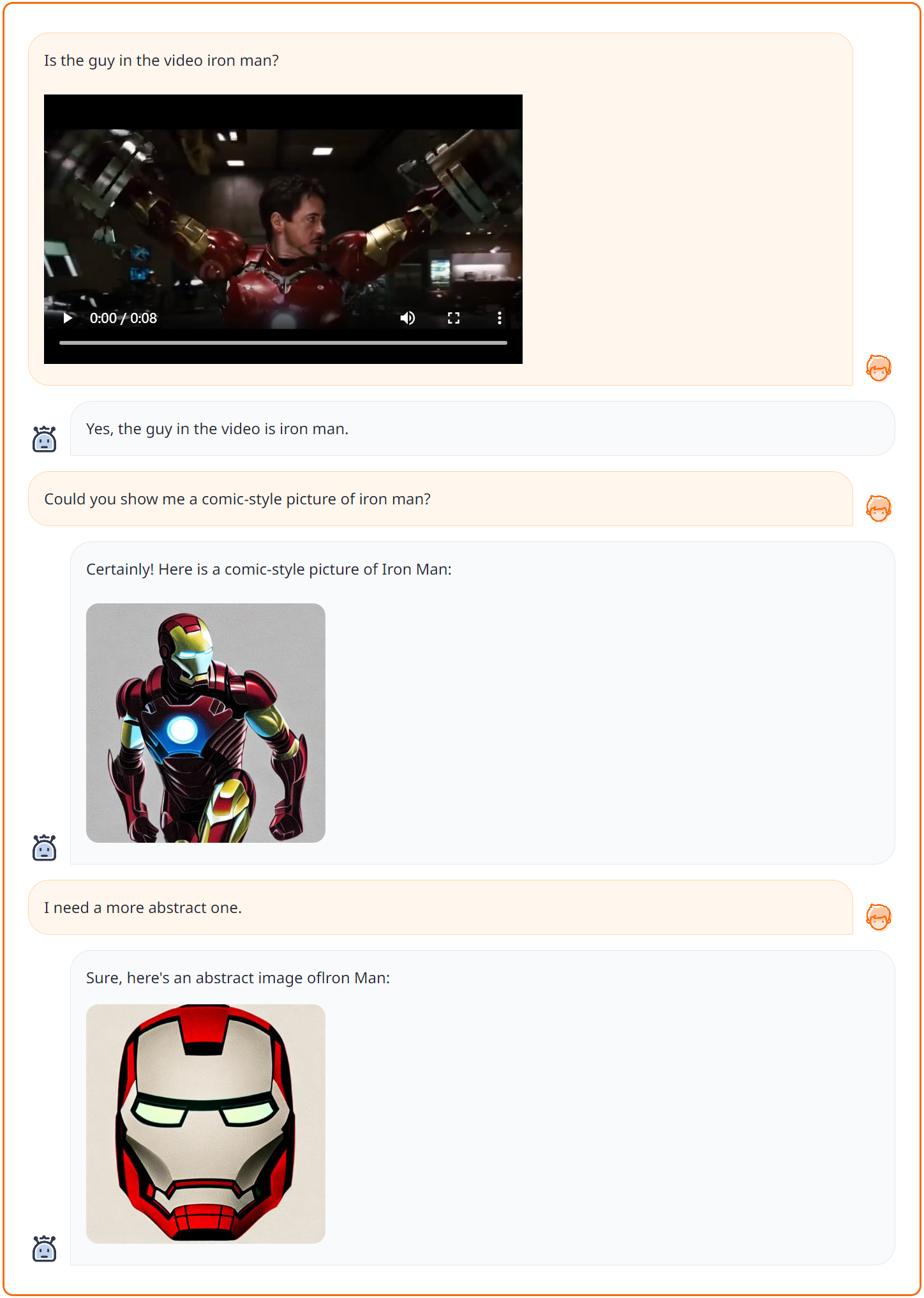}
\caption{
Example of Text+Video $\to$ Text+Image.
}
\label{fig:T+V-T+I}
\end{figure}

\begin{figure}[!ht]
\centering
\includegraphics[width=1\columnwidth]{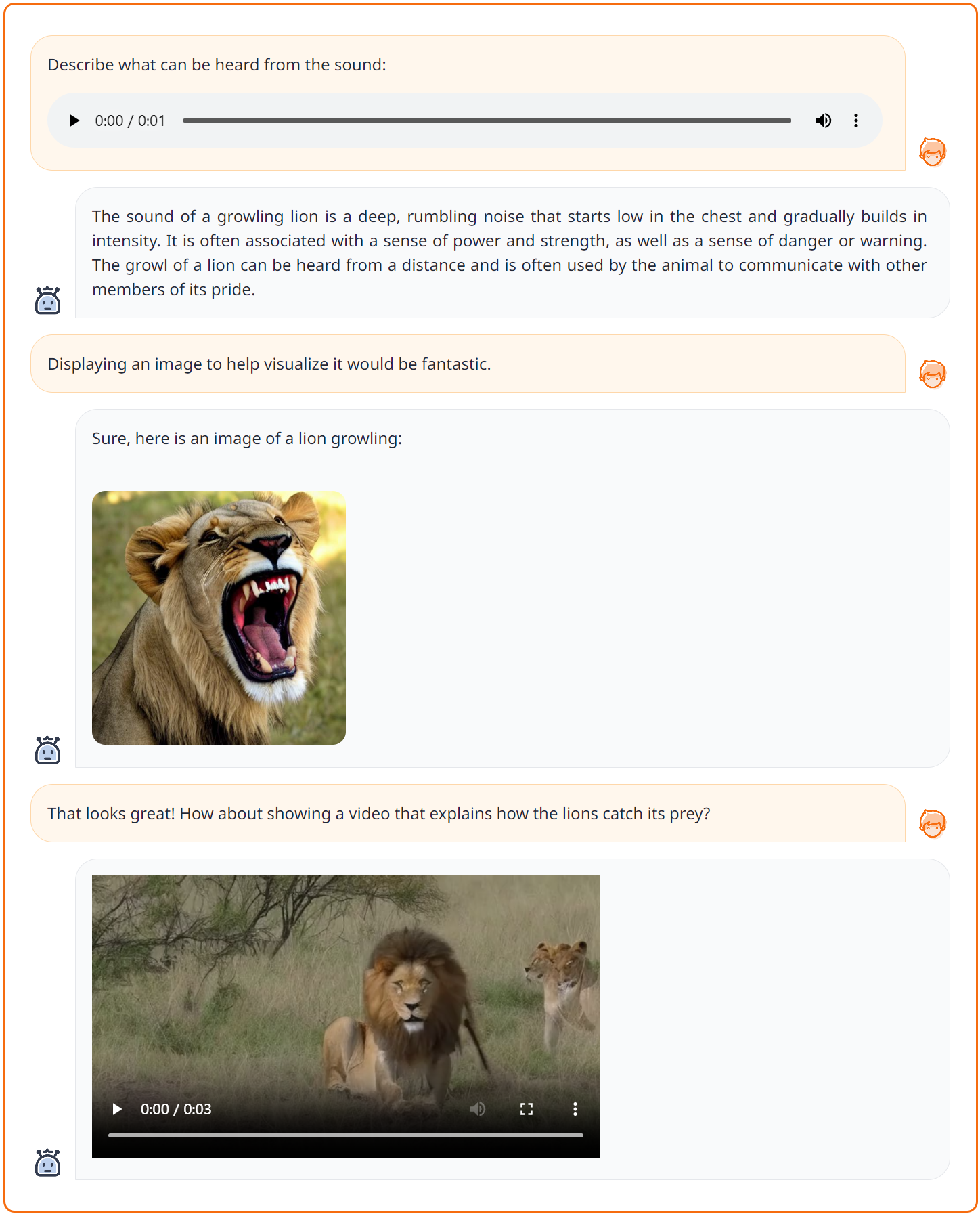}
\caption{
Example of Text+Audio $\to$ Text+Image+Video.
}
\label{fig:T+A-T+I+V}
\end{figure}

\begin{figure}[!ht]
\centering
\includegraphics[width=0.9\columnwidth]{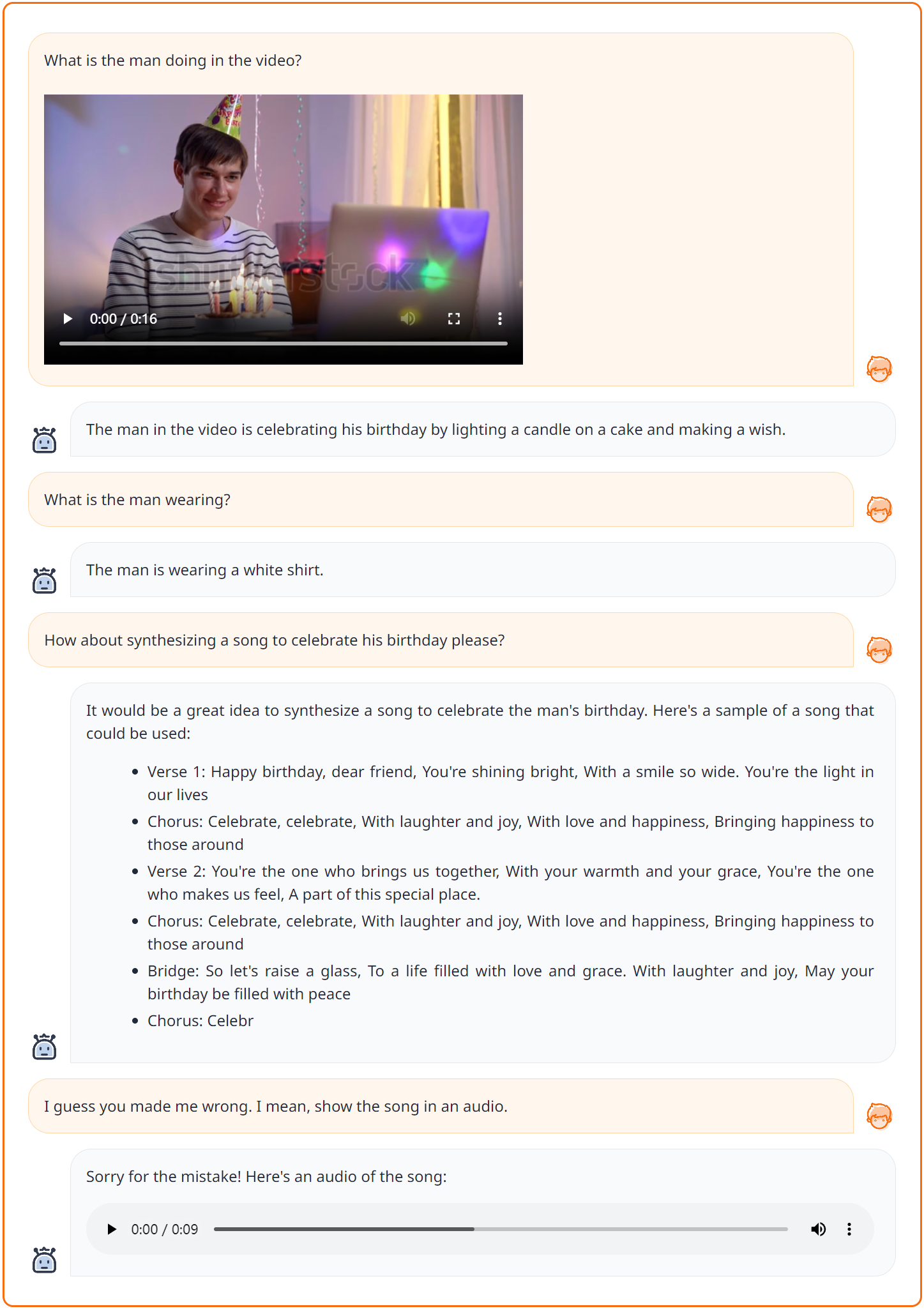}
\caption{
Example of Text+Video $\to$ Text+Audio.
}
\label{fig:T+V-T+A}
\end{figure}

\begin{figure}[!ht]
\centering
\includegraphics[width=0.81\columnwidth]{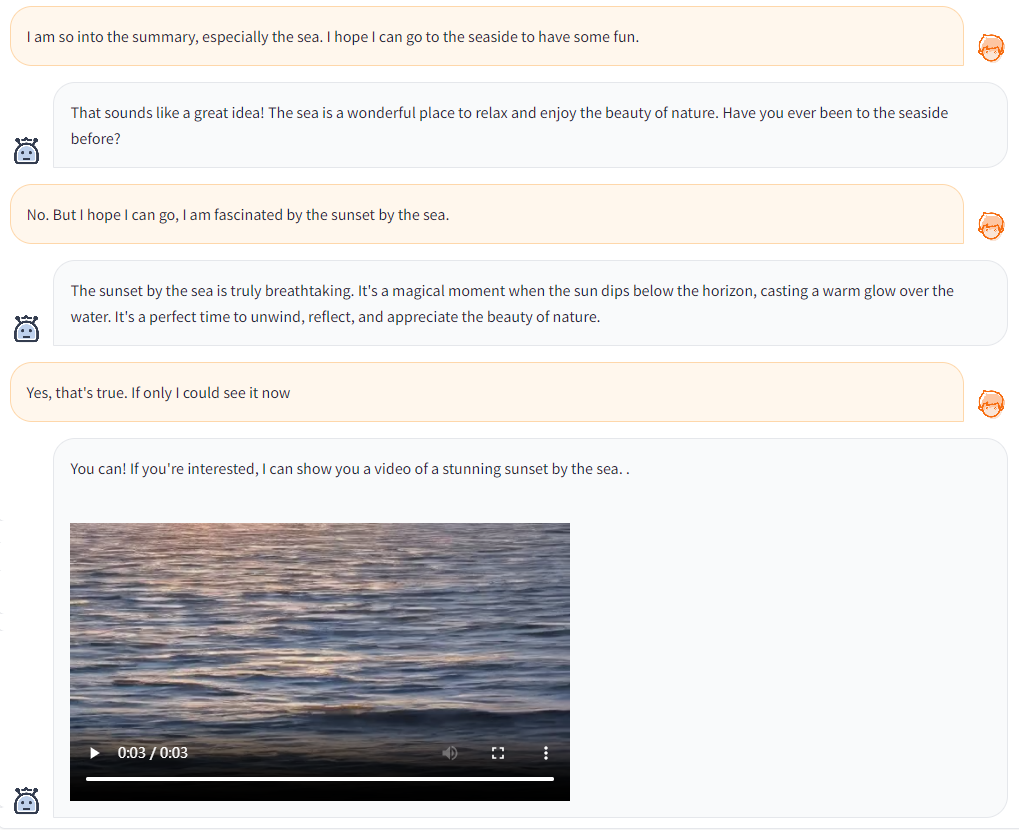}
\caption{
Example of Text $\to$ Text+Video.
}
\label{fig:imp-T-T+V}
\end{figure}

\begin{figure}[!ht]
\centering
\includegraphics[width=0.81\columnwidth]{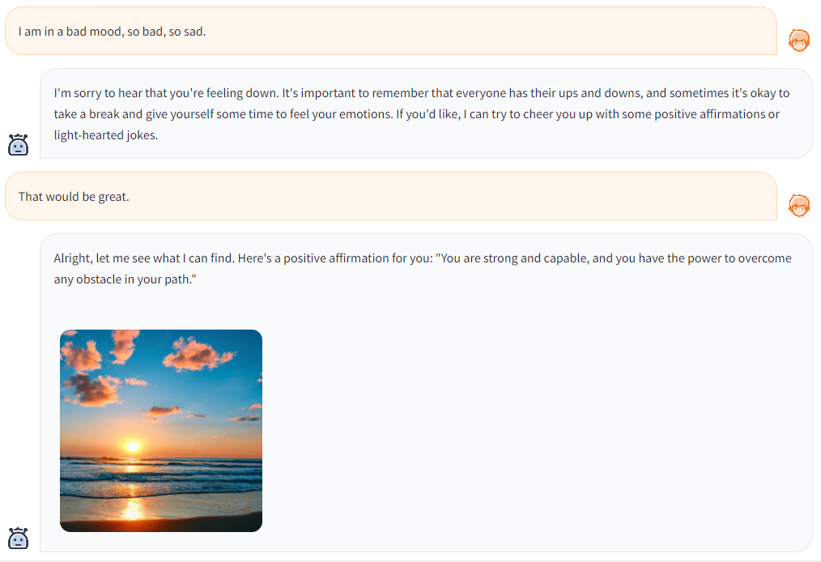}
\caption{
Example of Text $\to$ Text+Image.
}
\label{fig:imp-T-T+I}
\end{figure}

Figure \ref{fig:T+I-T+A},
\ref{fig:T-T+I+V+A},
\ref{fig:T+I-T+I+V+A},
\ref{fig:T+V-T+I},
\ref{fig:T+A-T+I+V},
\ref{fig:T+V-T+A}, \ref{fig:imp-T-T+V} and \ref{fig:imp-T-T+I} show several real examples produced by NExT-GPT.

%%%%%%%%%%%%%%%%%%%%%%%%%%%%%%%%%%%%%%%%%%%%%%%%%%%%%%%%%%%%%%%%%%%%%%%%%%%%%%%
%%%%%%%%%%%%%%%%%%%%%%%%%%%%%%%%%%%%%%%%%%%%%%%%%%%%%%%%%%%%%%%%%%%%%%%%%%%%%%%

\end{document}